\pdfoutput=1

\documentclass[11pt]{article}

\usepackage{acl}

\usepackage{times}
\usepackage{latexsym}
\usepackage{booktabs}
\usepackage{placeins}
\usepackage{enumitem}
\usepackage{amsmath}
\usepackage{amssymb}
\usepackage{multirow}
\usepackage{comment}
\usepackage{nicematrix}
\usepackage{pifont}
\usepackage{wasysym}
\usepackage{makecell}
\usepackage{ulem}
\usepackage{inconsolata}
\usepackage{balance}

\usepackage[T1]{fontenc}

\usepackage[utf8]{inputenc}

\usepackage{microtype}
\usepackage{graphicx}
\usepackage{soul}

\newcommand{\cmark}{\ding{51}}%
\newcommand{\xmark}{\ding{55}}%

\makeatletter
\DeclareRobustCommand{\iscircle}{\mathord{\mathpalette\is@circle\relax}}
\newcommand\is@circle[2]{%
  \begingroup
  \sbox\z@{\raisebox{\depth}{$\m@th#1\bigcirc$}}%
  \sbox\tw@{$#1\square$}%
  \resizebox{!}{\ht\tw@}{\usebox{\z@}}%
  \endgroup
}
\makeatother
\usepackage[dvipsnames, table]{xcolor}

\usepackage{circledsteps}
\newcommand\myCircled[2][]{\ifmmode
\Circled[fill color=black,inner color=white,#1]{\mathsf{#2}}
\else
\Circled[fill color=black,inner color=white,#1]{\sffamily#2}
\fi
}

\definecolor{blue}{HTML}{A0CBE8}
\definecolor{darkgreen}{HTML}{8CD17D}
\definecolor{lightgreen}{HTML}{BDF2B1}
\definecolor{orange}{HTML}{FFBE7D}
\definecolor{redbackground}{HTML}{FF9896}
\definecolor{greenbackground}{HTML}{98DF8A}

\newcommand{\ctext}[2]{%
  \begingroup
  \sethlcolor{#1}%
  \hl{#2}%
  \endgroup
}

\title{CHAMP: A Competition-level Dataset for Fine-Grained Analyses of LLMs' Mathematical Reasoning Capabilities}

\author{Yujun Mao\\ Boston University \\\texttt{amao@bu.edu} \And Yoon Kim\\MIT CSAIL\\\texttt{yoonkim@mit.edu} \\\\\url{https://yujunmao1.github.io/CHAMP/} \And Yilun Zhou\\Salesforce Research \\\texttt{yilun.zhou@salesforce.com}}

\begin{document}
\maketitle

\begin{abstract}
Recent large language models (LLMs) have shown indications of mathematical reasoning ability on challenging competition-level problems, especially with self-generated verbalizations of intermediate reasoning steps (i.e.,  chain-of-thought prompting). However, current evaluations mainly focus on the end-to-end final answer correctness, and it is unclear whether LLMs can make use of helpful side information such as problem-specific hints. In this paper, we propose a challenging benchmark dataset for enabling such analyses. The Concept and Hint-Annotated Math Problems (CHAMP) consists of high school math competition problems, annotated with \emph{concepts}, or general math facts, and \emph{hints}, or problem-specific tricks. These  annotations allow  us to explore the effects of additional information, such as relevant hints, misleading concepts, or related problems. This benchmark is difficult, with the best model only scoring 58.1\% in standard settings. With concepts and hints, performance sometimes improves, indicating that some models can make use of such side information. Furthermore, we annotate model-generated solutions for their correctness. Using this corpus, we find that models often arrive at the correct final answer through wrong reasoning steps. In addition, we test whether models are able to verify these solutions, and find that most models struggle. 

\end{abstract}

\section{Introduction}

\begin{figure}[t]
    \centering
    \includegraphics[width=0.95\columnwidth]{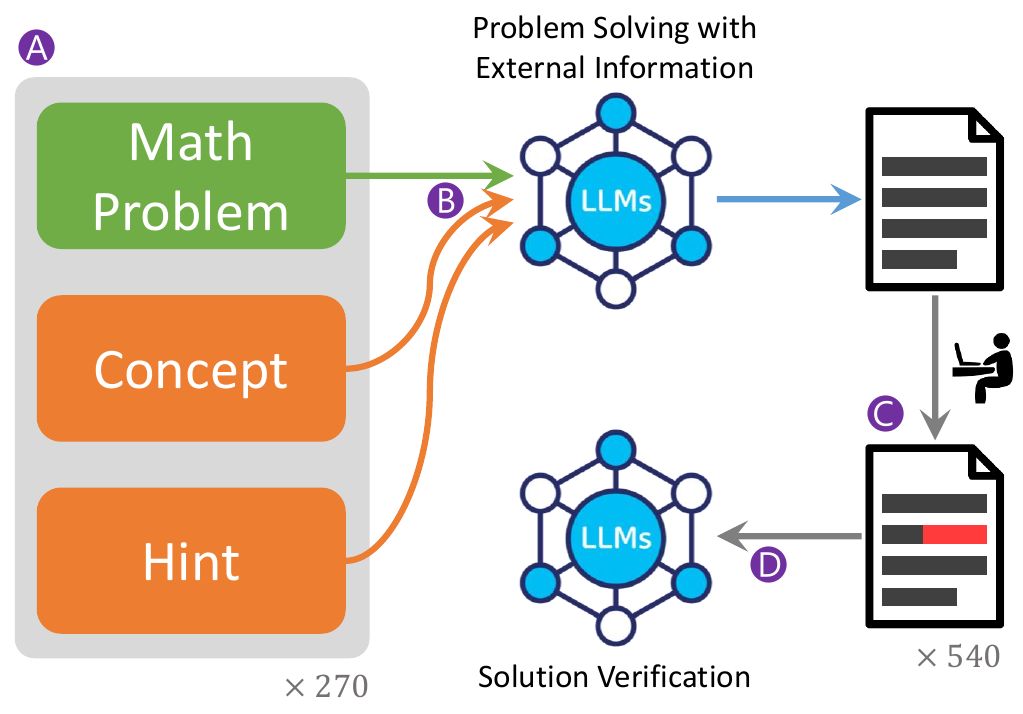}
    \caption{Overview of our dataset and experiment contribution. {\scriptsize \myCircled{A}}: We collect 270 challenging, high-school math competition problems (e.g., \textit{Find all positive integer solutions to the equation $x^3+3=4y(y+1)$}). For each problem, we write the relevant and helpful \textbf{C}oncepts (e.g., $a^3\pm b^3=(a+b)(a^2\pm ab+b^2)$), and \textbf{H}ints (e.g, \textit{Express $x^3$ as the product of two factors involving $y$}). {\scriptsize \myCircled{B}}: In our experiments, to investigate a model's ability to understand and use the additional C \& H information, we design 17 prompts to evaluate ten models: GPT-3.5 / 4 / 4 Turbo, PaLM 2 Medium, Llama 2 7B / 70B, Llama 3 8B / 70B, Mistral 7B and Mixtral 8x22B. {\scriptsize \myCircled{C}}: For each problem, we manually judge two model-generated solutions on their correctness, and further annotate the first wrong step of the reasoning (red highlights), if present. {\scriptsize \myCircled{D}}: This corpus thus serves as a novel dataset for benchmarking and evaluating the solution verification ability of LLMs.}
    \label{fig:1}
\end{figure}

Recent large language models (LLMs) have demonstrated impressive performance on many tasks that previously required specialized models or were thought to be out of reach of conventional neural networks. One such capability is mathematical reasoning: LLMs can often solve simple math problems  and make reasonable attempts at challenging, competition-level problems. In addition to model scaling \citep{kaplan2020scaling}, there are two key factors behind the progress: sophisticated prompting methods such as chain-of-thought \citep{wei2022chain, kojima2022large} and self-consistency \citep{wang2022self},  which provide useful heuristics for generating and selecting better reasoning paths; and access to calculators or code interpreters which offloads some of the symbolic computation to external tools \citep{gao2023pal, zhou2023solving}. However, a direction which remains less explored is how external concepts and hints impact  LLMs' reasoning abilities. This is difficult to address with existing datasets which typically only contain  problem statements and their solutions, and do not provide annotated concepts or hints that would be helpful for the problem at hand.

To enable such analyses, we introduce the Concept and Hint-Annotated Math Problems (CHAMP) dataset, which consists of 270 diverse high school competition-level math problems (Fig.~\ref{fig:1}). In addition to problem statements and full solutions, we annotate each problem with two key pieces of information: concepts and hints. Concepts are general math theorems or formulas, while hints are problem-specific tricks or strategies. The design of CHAMP enables previously under-explored evaluations of multi-step reasoning abilities of LLMs. For example, can LLMs make use of these concepts and hints? How should these be provided to the LLM? Could a model infer useful information from studying sample problems using the same concepts? What happens if the model is provided with irrelevant/misleading concepts? 

Using this dataset, we design 17 different prompts, and evaluate various proprietary and open-source models including GPT-3.5 / 4 / 4 Turbo \citep{OpenAI2023GPT4TR}, PaLM 2 Medium \citep{anil2023palm}\footnote{The API access for PaLM 2 Large was not publicly available at the time of experiments.}, Llama 2 7B / 70B \citep{touvron2023llama}, Llama 3 8B / 70B \citep{meta2024introducing}, Mistral 7B \citep{jiang2023mistral} and Mixtral 8x22B \citep{AI_2024}.  While we observe a diverse range of behaviors across different models and prompts, we find the accuracy of the best setting to be 67.0\% only (measured by the final answer correctness, ignoring any possible errors in intermediate reasoning steps), and gains from the additional concept and hint information vary across models. The results indicate a large room for improvement with competition-level math for LLMs, and moreover highlight the utility of CHAMP for developing and benchmarking future models. 

For each problem, we further analyze solutions generated by these models by manually annotating the first wrong step in the reasoning process, such as an arithmetic error or question misunderstanding, or validating that the solution is fully correct. This annotation serves two purposes. First, it concretely identifies how much the final answer accuracy (which is the predominant practice for math problem evaluations) over-estimates a model's ability to generate fully correct solutions: indeed, we find that in many instances, a model gets the answer ``right'' despite generating wrong reasoning steps, indicating that these models are potentially relying on shortcut heuristics. Second, we can evaluate the verification ability of any model, i.e., how well it can reason about a given solution and identify any errors, where we find that most models struggle. The above evaluations suggest key deficiencies in current LLMs and highlight the value of these annotations as an additional benchmarking resource.

In summary, we evaluate four model capabilities using our dataset: generating correct final answer, generating correct full solution, using the contextual C \& H information, and verifying a given solution. Our findings uncover new strengths and limitations of current models, and give directions for future work on improving them. 

\section{Background and Related Work}
\paragraph{Math datasets and benchmarks.}
Large language models have seen significant improvement in understanding and solving math problems, with GPT-4 \citep{OpenAI2023GPT4TR} being able to tackle most math problems that require grade-level knowledge and direct applications of formulas, even for problems with diverse formats and wordings \citep{cobbe2021training}. Nonetheless, they still struggle with competition-level problems, such as those found in the MATH dataset \citep{hendrycks2021measuring}. Competition-level problems---for which applications of formulas are not straightforward---are therefore the focus of our CHAMP dataset. 

A key distinction of CHAMP compared to other math datasets 
is the information associated with each problem. In addition to the problem text and its solution---which are common components of such datasets---we annotate relevant concepts and hints and further label them on each solution step, with problems relating to each other via common math concepts (e.g., Fermat's little theorem). In this way, CHAMP enables fine-grained evaluations of mathematical problem solving abilities of LLMs that are not possible with other datasets, for example allowing for different types of additional information to be made available in the context through prompting.

We observe that many techniques seek to improve an LLM's mathematical reasoning ability, such as encouraging chain-of-thought generation \citep{wei2022chain}, selecting a final result from multiple sampled outputs \citep{wang2022self}, and using external tools such as a calculator or Python interpreter \citep{gao2023pal} to eliminate arithmetic errors. These directions   can be combined with our experimental setup.

\paragraph{Solution verification ability of LLMs.} Another distinguishing factor of our dataset is the first wrong step annotations on model solutions, which enables more fine-grained model analyses and, more importantly, evaluations of how well models can \textit{verify} a given answer. 

There have been recent attempts at crafting such datasets. For example, \citet{lightman2023let} collected PRM800K, containing 800K steps of 75K solutions to 12K problems in the MATH dataset \citep{hendrycks2021measuring}, with each step labeled as correct, incorrect or neutral. \citet{chen2023felm} curated FELM, a factuality benchmark, including annotations of solutions to 208 GSM8K \citep{cobbe2021training} and 194 MATH problems. Compared to CHAMP, where annotations are made exclusively by the paper authors, both PRM800K and FELM are labeled via crowdsourcing. Moreover, solutions in PRM800K are selected to maximally confuse a reward model being developed in the project, while FELM uses only GPT-3.5 as the solution generator. In contrast, our 540 annotated solutions are generated by a mix of GPT-3.5, 4, 4 Turbo and PaLM 2 Medium, each with varying capabilities. 

\paragraph{Roles of contexts.} Our dataset and experiments are similar in spirit to works that explore how well LLMs understand different contexts, which have yielded surprising findings. For example, models can be insensitive to label correctness \citep{min2022rethinking} but sensitive to label distribution \citep{zhao2021calibrate} and exemplar ordering \citep{lu2021fantastically}. \citet{mckenzie2023inverse} find that larger LLMs resist absorbing context information inconsistent with world knowledge acquired during training (e.g., redefining $\pi=432$). Similarly, \citet{wu2023reasoning} find that LLMs perform worse in atypical setups for common tasks (e.g., base-9 integer addition). With CHAMP, we can explore how different information supplied in various ways affect LLMs' behavior.


\section{The CHAMP Dataset}
\label{sec:dataset}
This section describes the dataset structure and construction. Due to the high level of math expertise required, the dataset curation is carried out exclusively by the paper authors.

\begin{table}[!t]
    \centering
    \resizebox{\columnwidth}{!}{
    \begin{tabular}{p{9cm}}\toprule\midrule
        Problem ID: P\_Inequality\_36\newline 
         Problem: For non-negative $a, b, c, d$, what is the smallest value of $\sqrt{(a+c)(b+d)}-\sqrt{ab}-\sqrt{cd}-1$? \\\midrule
         \textbf{C}oncepts and \textbf{H}ints: \begin{enumerate}[leftmargin=2em, topsep=0pt, itemsep=0pt, parsep=0pt]
         \item[H1.] Compare $\sqrt{(a+c)(b+d)}$ with $\sqrt{ab}+\sqrt{cd}$ by squaring both terms.
         \item[C1.] $(x\pm y)^2=x^2\pm 2xy+y^2$.
         \item[C2.] For non-negative $x, y$, we have $(x+y)\geq 2\sqrt{xy}$ and $(x+y)/2\geq \sqrt{xy}$, with equality if and only if $x=y$.
         \item[C3.] For non-negative $x, y$, $\sqrt{x}\geq \sqrt{y}$ if and only if $x\geq y$.
         \end{enumerate}
         \\\midrule
         Answer: $-1$\\
         Solution Step: 
         \begin{enumerate}[leftmargin=*, topsep=0pt, itemsep=0pt, parsep=0pt]
         \item We have $\sqrt{(a+c)(b+d)}^2=(a+c)(b+d)=ab+ad+bc+cd$. [H1]
         \item We have $(\sqrt{ab}+\sqrt{cd})^2=ab+cd+2\sqrt{abcd}$. [H1, C1]
         \item Thus, $\sqrt{(a+c)(b+d)}^2-(\sqrt{ab}+\sqrt{cd})^2=ad+bc-2\sqrt{abcd}$, which is non-negative because $ad+bc\geq 2\sqrt{abcd}$. [H1, C2]
         \item Thus, $\sqrt{(a+c)(b+d)}^2\geq {sqrt(ab)+\sqrt(cd)}^2$.
         \item Since $a, b, c, d$ are all non-negative, we have $\sqrt{(a+c)(b+d)}\geq \sqrt{ab}+\sqrt{cd}$. [C3]
         \item So the smallest value of $\sqrt{(a+c)(b+d)}-\sqrt{ab}-\sqrt{cd}-1$ is $-1$, achieved when $a=b=c=d$.
         \end{enumerate}\\ \bottomrule
    \end{tabular}
    }
    \vspace{-2mm}
    \caption{A sample from the CHAMP dataset, which shows the problem (top), the concepts and hints (middle), and the full solution (bottom).}
    \vspace{-2mm}
    \label{tab:sample-problem-main}
\end{table}

\paragraph{Problems.} We select problems from 
the book \textit{Problem-Solving Strategies} by \citet{engel2008problem}, a classic piece of material for high-school math competitions. All problems require specific tricks or creative strategies, rather than routine knowledge applications. We require problems to have final check-able answers for easy evaluation, and thus rewrite proof problems where possible (e.g., ``\text{Prove $f(x) \geq 1$ for $x \geq 0$}'' is transformed to ``\text{What is the smallest value of $f(x)$ for $x\geq 0$}''). A total of 270 problems span five categories: number theory (80), polynomial (50), sequence (50), inequality (50) and combinatorics (40). We make some adaptations to lessen the impact of ``trivial limitations'' of LLMs, such as weakness in precise arithmetics. See App. \ref{app:category} for further details.

For each problem, we manually verify and write out the full detailed step-wise solution in natural language, as the solution manual often skips steps and occasionally contains typographical errors. We also provide an explicit final answer that can be checked against.

\begin{figure*}
    \centering
    \includegraphics[width=\textwidth]{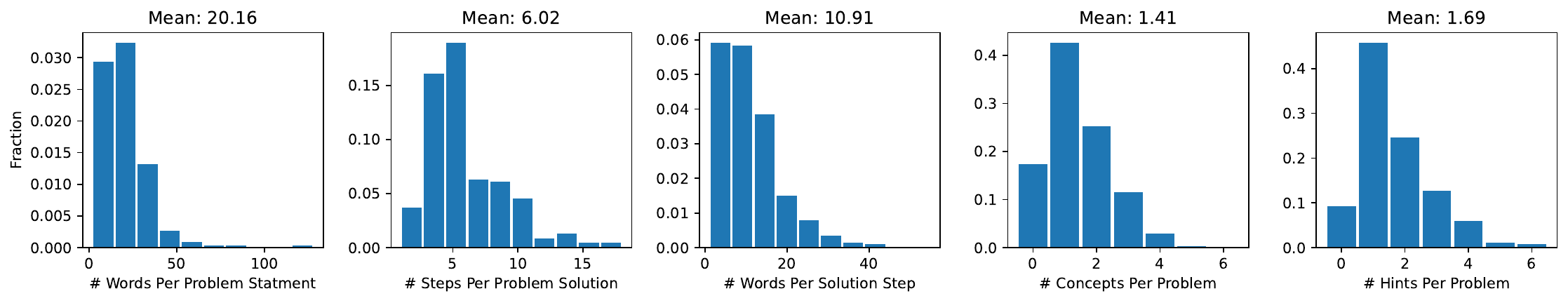}
 \vspace{-5mm}
    \caption{The distribution of dataset statistics in CHAMP. Problems in CHAMP require a nontrivial number of reasoning steps (6.0 on average). Each problem is linked to an average of 1.4 concepts and 1.7 hints.}
\vspace{-3mm}
    \label{fig:stats}
\end{figure*}

\paragraph{Concepts and hints.}
We additionally annotate relevant \textit{concepts} and \textit{hints}, which provide helpful information to solve a problem. For the purposes of this paper, we define concepts to mean general math knowledge, such as an equation or a theorem, for example, ``$x^2-y^2=(x+y)(x-y)$''. We define hints, on the other hand, to be problem-specific tricks or strategies, such as ``\text{Add 1 to both sides.}'' (to prepare for further simplification). These concepts and hints are also labeled to the relevant solution steps in the ground truth solution.

Each concept is additionally annotated with three metadata fields: the \ul{\textit{category}}, such as ``number theory'' or ``polynomial''; the \ul{\textit{name}}, such as ``difference of squares formula'' for the concept $x^2-y^2=(x+y)(x-y)$; and the \ul{\textit{parent concept}}, which is a general form of the concept. For example, both $(x+y)^2=x^2+2xy+y^2$ and $(x+y)^3=x^3+3x^2y+3xy^2+y^3$ have the parent concept of the binomial theorem, which states $(x+y)^n=\sum_{k=0}^n{n\choose k}x^{(n-k)}y^k$. While every concept is assigned a category, not all concepts have names or parent concepts. One CHAMP problem with its solution and concept \& hint annotation is shown in Tab.~\ref{tab:sample-problem-main}; additional examples are given in Tab.~\ref{tab:sample-problem} of App.~\ref{app:dataset}.


\paragraph{First wrong step (FWS) annotations.}

For every problem, we further analyze the LLM-generated solutions to identify mistakes in reasoning steps. Specifically, we randomly assign a ``generator model,'' one of GPT-3.5, GPT-4, GPT-4 Turbo and PaLM 2 M, to each problem, stratified by the problem's category (i.e., number theory, polynomial, etc.), and collect the model-generated solutions for two prompts: one with the problem statement only (the ``No C + w/o H'' prompt in Tab.~\ref{tab:result-table} of Sec.~\ref{sec:model-eval}) and another with the problem statement supplemented by the list of relevant concepts and hints (the ``Direct + w/H'' in Tab.~\ref{tab:result-table} of Sec.~\ref{sec:model-eval}). We assess whether the \textit{full solution} (including the reasoning steps) is correct, and if not, annotate the occurrence of the first wrong step (FWS) as a span in the text. The FWS refers to an objective mistake, such as an arithmetic error  or question misunderstanding, and does not contain subjective assessments, such as a seemingly implausible strategy to the problem.

\paragraph{Dataset statistics.}
We collect 270 problems, 54 concepts, 330 hints and 540 FWS annotations. Every problem has at least 1 concept or hint, with an average of 1.4 concepts and 1.7 hints, and an average of 6.0 solution steps. Each problem statement has an average of 20.2 words, and each solution step 10.9 words. Fig.~\ref{fig:stats} plots the distribution histograms of these statistics.

\section{Experiment 1: Problem Solving}
\subsection{Experimental Setup}
\label{sec:setup}
We evaluate ten models: GPT-3.5 Turbo (16k context version), GPT-4, GPT-4 Turbo, PaLM 2 Medium, Llama 2 7B, Llama 2 70B, Llama 3 8B, Llama 3 70B, Mistral 7B and Mixtral 8x22B (exact model versions in Tab.~\ref{tab:result-table}). 
We set temperature to 0 for all experiments, and allow at least 1024 tokens to be generated for each model depending on the specific context window size, which is more than enough to output the correct reasoning, with unfinished generations treated as incorrect. Inference for proprietary models is done via the respective API, and that for open-source models is done via the \url{together.ai} API.


\paragraph{Baseline.} Sometimes models can output correct final answers with incorrect reasoning, especially when those answer values appear often in the training corpus. To contextualize the final answer accuracy metrics, we construct the majority guess baseline as follows. For each of the four answer types in our dataset---numeric answers (e.g., 0), expression answers (e.g., $n$), yes/no answers (e.g., no) and enumeration answers (e.g., $x=1$ or $x=-2$)---the baseline guesses the most frequently appearing answer values in the ground truth. This baseline accuracy is 33.0\%. See Tab. \ref{tab:baseline} of App. \ref{app:dataset} for more details.

\begin{table*}[!htb]
    \centering
    \resizebox{\textwidth}{!}{
    \begin{NiceTabular}{r|cccc|rcccccccc}
    \CodeBefore
      \rectanglecolor{blue}{3-2}{22-5}
      \rectanglecolor{lightgreen}{3-7}{3-12}
      \rectanglecolor{darkgreen}{4-7}{4-12}
      \rectanglecolor{lightgreen}{5-7}{5-12}
      \rectanglecolor{darkgreen}{6-7}{6-12}
      \rectanglecolor{lightgreen}{7-7}{7-12}
      \rectanglecolor{darkgreen}{8-7}{8-12}
      \rectanglecolor{lightgreen}{9-7}{9-12}
      \rectanglecolor{darkgreen}{10-7}{10-12}
      \rectanglecolor{lightgreen}{11-7}{11-12}
      \rectanglecolor{darkgreen}{12-7}{12-12}
      \rectanglecolor{lightgreen}{13-7}{13-12}
      \rectanglecolor{darkgreen}{14-7}{14-12}
      \rectanglecolor{lightgreen}{15-7}{15-12}
      \rectanglecolor{darkgreen}{16-7}{16-12}
      \rectanglecolor{lightgreen}{17-7}{17-12}
      \rectanglecolor{darkgreen}{18-7}{18-12}
      \rectanglecolor{lightgreen}{19-7}{19-12}
      \rectanglecolor{darkgreen}{20-7}{20-12}
      \rectanglecolor{lightgreen}{21-7}{21-12}
      \rectanglecolor{darkgreen}{22-7}{22-12}
      \rectanglecolor{orange}{3-13}{22-13}
    \Body
    \toprule
    & \multicolumn{4}{c}{Standard Evaluations} & \multicolumn{8}{c}{CHAMP-Enabled Evaluations (7 Ways to Provide Concepts With or Without Hints)}\\
    Model & 0-Shot & 5-Shot & 1/3 Soln & 2/3 Soln & & 1. No C & 2. Direct & 3. Root & 4. Name & 5. Example & 6. Problem & 7. Misleading\\\midrule\midrule
    GPT-3.5 & \multirow{2}{*}{28.5} & \multirow{2}{*}{34.8} & \multirow{2}{*}{33.7} & \multirow{2}{*}{40.7} & w/o H & 28.5 & 28.5 & 28.1 & \textbf{33.0}&30.4&30.0&27.8\\
    \scriptsize{gpt-3.5-turbo-16k-0613} & & & & &w/ H & 31.9 & 33.0&33.0&31.9&31.9&\textbf{34.4} & 31.5  \\\midrule
    GPT-4 & \multirow{2}{*}{41.9} & \multirow{2}{*}{38.1} & \multirow{2}{*}{53.7} & \multirow{2}{*}{65.6} & w/o H & 41.9&\textbf{43.0}&42.6&40.0&42.2&43.7&39.3\\
    \scriptsize{gpt-4-0613} &&&&& w/ H & 51.9&\textbf{53.0}&49.6&49.3&52.2&51.1&48.9  \\\midrule
    GPT-4 Turbo & \multirow{2}{*}{58.1} & \multirow{2}{*}{53.0} & \multirow{2}{*}{61.5} & \multirow{2}{*}{67.0} & w/o H & \textbf{58.1}&57.0&55.6&51.1&55.9&55.2&51.9\\ 
    \scriptsize{gpt-4-1106-preview} &&&&& w/ H& 62.2&\textbf{65.9}&64.8&63.0&63.3&64.4&55.6  \\\midrule
    PaLM 2 Medium & \multirow{2}{*}{14.1} & \multirow{2}{*}{17.4} & \multirow{2}{*}{15.9} & \multirow{2}{*}{23.3} & w/o H & 14.1&15.6&15.6&14.1&\textbf{17.0}&16.7&19.3 \\ 
    \scriptsize{chat-bison-001} &&&&& w/ H & 14.4&15.2&15.2&18.1&16.7&\textbf{20.7}&19.3 \\\midrule
    Llama 2 7B & \multirow{2}{*}{\phantom{0}8.5} & \multirow{2}{*}{\phantom{0}9.6} & \multirow{2}{*}{\phantom{0}7.4} & \multirow{2}{*}{10.7} & w/o H & \phantom{0}8.5 & \phantom{0}7.8 & \phantom{0}8.1 & \phantom{0}6.7 & \phantom{0}9.3 & \textbf{10.0} & \phantom{0}9.3 \\ 
    \scriptsize{Llama-2-7b-chat-hf} &&&&& w/ H & \phantom{0}9.6 & \phantom{0}7.4 & \phantom{0}9.3 & \phantom{0}7.4 & 10.0 & \textbf{11.1} & \phantom{0}8.9 \\\midrule
    Llama 2 70B & \multirow{2}{*}{11.9} & \multirow{2}{*}{13.0} & \multirow{2}{*}{14.8} & \multirow{2}{*}{21.5} & w/o H & 11.9 & 13.3 & 11.1 & \textbf{14.1} & 13.7 & 13.3 & 12.6 \\ 
    \scriptsize{Llama-2-70b-chat-hf} &&&&& w/ H & 15.6 & 15.6 & 15.9 & 15.9 & \textbf{16.7} & 15.6 & 16.3 \\\midrule
    Llama 3 8B & \multirow{2}{*}{20.7} & \multirow{2}{*}{22.2} & \multirow{2}{*}{25.2} & \multirow{2}{*}{35.9} & w/o H & 21.1 & 23.0 & 24.1 & \textbf{24.8} & 20.0 & \textbf{24.8} & 20.7 \\ 
    \scriptsize{Meta-Llama-3-8B-Instruct} &&&&& w/ H & 23.7 & 25.6 & 25.6 & \textbf{29.3} & 26.3 & 25.9 & 21.5 \\\midrule
    Llama 3 70B & \multirow{2}{*}{37.8} & \multirow{2}{*}{35.9} & \multirow{2}{*}{47.0} & \multirow{2}{*}{59.6} & w/o H & 37.8 & \textbf{40.0} & 37.0 & \textbf{40.0} & 38.1 & 37.0 & 35.6 \\ 
    \scriptsize{Meta-Llama-3-70B-Instruct} &&&&& w/ H & 47.0 & 49.6 & 48.9 & 47.4 & \textbf{50.0} & 48.9 & 42.2 \\\midrule
    Mistral 7B & \multirow{2}{*}{20.7} & \multirow{2}{*}{16.3} & \multirow{2}{*}{18.1} & \multirow{2}{*}{24.8} & w/o H & \textbf{20.7} & 20.4 & 20.0 & \textbf{20.7} & 19.3 & 19.6 & 18.1 \\ 
    \scriptsize{Mistral-7B-Instruct-v0.3} &&&&& w/ H & 18.1 & 19.3 & 17.0 & 20.0 & 18.9 & \textbf{22.2} & 16.7 \\\midrule
    Mixtral 8x22B & \multirow{2}{*}{36.7} & \multirow{2}{*}{36.7} & \multirow{2}{*}{47.0} & \multirow{2}{*}{60.7} & w/o H & 36.7 & 38.1 & \textbf{39.3} & 32.6 & 34.1 & 38.5 & 32.2 \\ 
    \scriptsize{Mixtral-8x22B-Instruct-v0.1} &&&&& w/ H & 47.4 & 48.5 & 49.6 & 47.0 & 46.3 & \textbf{50.0} & 45.2 \\\midrule
    \multicolumn{13}{c}{Majority Guess Baseline: 33.0}\\\bottomrule
    \end{NiceTabular}
    }
\vspace{-2mm}
    
    \caption{Final answer accuracy (in percentage) with the  different prompt settings.}
 \vspace{-3mm}
    
    \label{tab:result-table}
\end{table*}

\paragraph{Automatic evaluation.} 
As is the case with many generation tasks, the correct final answer can be expressed in multiple ways:
``no solution'' is equivalent to ``none'', ``unsolvable'' is equivalent to ``impossible to solve'', etc. Therefore a simple criteria-based exact or sub-string match to a manually constructed set of valid answers is prone to false negatives. We thus propose to use GPT-4 as an automatic grader, and use a three-stage procedure for the solution generation and evaluation. First, we prompt the model for its step-by-step solution. Then, we ask the model to produce a one-sentence summarization of its answer. Finally, we use GPT-4 to grade the answer summary, given the ground truth final answer, which essentially checks for semantic equivalence between the two. The prompts are listed in Tab.~\ref{tab:prompt-problem-solving} and \ref{tab:prompt-grading} of App.~\ref{app:prompts}. 

To assess the validity of this automatic evaluation procedure, we manually checked 500 examples for (1) whether the one-sentence summarization was correct and (2) whether GPT-4 was able to correctly grade the summarized solution given the ground truth final answer. While not perfect, we found GPT-4 to be quite good at this, with the accuracies on both tasks being $\geq 97\%$.

\subsection{Model Analyses}
\label{sec:model-eval}

Our experiments are aimed at evaluating four different aspects: raw model performance, effectiveness of different ways to provide concepts, relative importance of concepts and hints, and impact of irrelevant (i.e., misleading) concepts. The quantitative results are summarized in Tab.~\ref{tab:result-table}.

\paragraph{Model performance.}

We first study the model performance with both zero-shot \citep{kojima2022large} and few-shot \citep{wei2022chain} chain-of-thought prompting. Following the experiments by \citep{hendrycks2021measuring}, we also study models' performance when given partial solution (1/3 and 2/3 of solution steps) under the zero-shot setup. The prompts are listed in Tab.~\ref{tab:prompt-0shot}-\ref{tab:prompt-partial} of App.~\ref{app:prompts}.


The \ctext{blue}{blue cells} of Tab.~\ref{tab:result-table} summarize these results, with some full model-generated solutions presented in Tab.~\ref{tab:full-solution-1} and \ref{tab:full-solution-2} of App.~\ref{app:fws-solution}. Generally, larger and more recent models perform better than their smaller and earlier versions, and partial solutions are mostly helpful in guiding the model to correct solutions, largely consistent with the findings by \citet{hendrycks2021measuring}. Overall, the best performing models, GPT-4 and GPT-4 Turbo, are still proprietary ones and there is a gap to close for (even the latest) open-source models. In addition, five-shot prompting is often not beneficial, suggesting that such instruction-tuned models, especially high-performing ones, may not need in-context exemplars to activate the ``problem-solving mode,'' as long as the instruction is sufficiently clear.


\paragraph{Concept provision method.} As concepts are general facts or formulas in math, they are likely already learned by the model during pre-training. However, different ways to provide the knowledge in the context may affect how well the model can understand and use it. We designed six concept provision approaches, each in two versions where the hint is withheld or provided, corresponding to the \ctext{lightgreen}{light green cells} and \ctext{darkgreen}{dark green cells} of Tab.~\ref{tab:result-table}: 
\begin{enumerate}[leftmargin=*, parsep=-3pt]
    \item Prompt with {no concepts} (No C).
    \item {Directly} provide the concept in the prompt (Direct). 
    \item Provide the {root} concept up the parent-child chain (i.e., most general form) in the prompt (Root).
    \item Ask the model to retrieve the concept by its {name} in a separate round of conversation (Name).
    \item Ask the model to provide an {example} for the concept in a separate round of conversation (Example).
    \item Provide a sample {problem} that uses the concept and its step-by-step solution (Problem).
\end{enumerate}
The specific prompts are listed in Tab.~\ref{tab:prompt-noC}-\ref{tab:prompt-problemC} of App.~\ref{app:prompts}. The best performing concept provision method for each model with and without hints is \textbf{bolded}. No single method performs the best across the board. Furthermore, concepts may sometimes even be detrimental, potentially because they contradict with the model's ``initial planned approach.''


\paragraph{Importance of hints.} Compared to concepts, hints are arguably more valuable as they often require creativity from the test-takers. The performance contrast without and with the hint under each prompt are shown by the \ctext{lightgreen}{light green rows} vs. \ctext{darkgreen}{dark green rows} of Tab.~\ref{tab:result-table}. While providing hints helps, the performance increase is correlated with the model's ``base capability'': GPT-4, 4 Turbo, Llama 3 70B and Mixtral 8x22B, which have the best zero-shot accuracy, are able to score 10\% higher on average with hints, while the accuracy increase is much less notable for other models.



\paragraph{Impact of misleading concepts.}

How well could a model deal with misleading concepts? In this experiment, for each useful concept, we replace it with a random one of the same category (to ensure maximal confusion) but not on the path to its root concept (to avoid accidentally providing useful information).\footnote{We do not experiment with misleading hints, as they would appear nonsensical due to problem-specificity.} The results are summarized in the \ctext{orange}{orange cells} of Tab.~\ref{tab:result-table}. Compared to the ``No C'' setup, misleading concepts have different impacts on different models: while most of the models show slight drop of accuracy with misleading concepts, GPT-4 Turbo suffers the most, with over 10\% relative decrease of accuracy compared with the ``No C'' setup. On the other hand, PaLM 2 Medium and both Llama 2 models even show some improvement, indicating that they are unlikely to understand and act on provided (mis)information.

\subsection{Full Solution Accuracy}
The above analyses are based on an important assumption: that the final answer accuracy is a faithful reflection of the model's mathematical ability. However, focusing on final answer alone could inflate the models' performance as incorrect reasoning could lead to correct final answers, especially for questions with yes/no answers (see examples in Tab.~\ref{tab:full-solution-1} and \ref{tab:full-solution-2} of App.~\ref{app:fws-solution}). As a result, we proceed to examine the full solutions generated by four models: GPT-3.5, GPT-4, GPT-4 Turbo and PaLM 2 Medium, based on the first wrong step (FWS) annotation.


\begin{table}[!t]
    \centering
    \resizebox{\columnwidth}{!}{
    \begin{NiceTabular}{r|rr|rr}\toprule
        & \multicolumn{2}{c|}{Problem Only} & \multicolumn{2}{c}{Problem + C \& H}\\
        & Final Ans & Full Soln & Final Ans & Full Soln\\\midrule
        GPT-3.5 & 34.3\% & 6.0\% & 32.8\% & 16.4\% \\
        GPT-4 & 54.4\% & 17.6\% & 52.9\% & 33.8\%\\
        GPT-4 T & 56.7\% & 22.4\% & 68.7\% & 44.8\%\\
        PaLM 2 M & 19.1\% & 1.5\% & 14.7\% & 0.0\%\\\bottomrule
    \end{NiceTabular}
    }
\vspace{-2mm}
    
    \caption{Final answer vs. full solution accuracy for four models under two prompts.}
\vspace{-3mm}
    
    \label{tab:full-solution-accuracy}
\end{table}

Tab.~\ref{tab:full-solution-accuracy} displays the final answer accuracy (FAA) and full solution accuracy (FSA) of model outputs from the 0-shot problem-only prompt and the problem + concept \& hint list prompt.\footnote{Note that the FAA statistics in Tab.~\ref{tab:full-solution-accuracy} are based on model-generated solutions of 25\% sampled problems for each model, and hence different from those in Tab.~\ref{tab:result-table}.} Full solution accuracy (FSA) is significantly lower than final answer accuracy (FAA), suggesting that the latter is indeed an inflated measurement of the models' true reasoning ability. As an extreme case, PaLM 2 M produces only one fully correct solution out of 136 attempts on 68 problems, despite still achieving 14.7\% FAA. Given that almost all benchmarks \citep[e.g.][]{cobbe2021training, hendrycks2021measuring} are based on FAA, true model performance may not be as high as previous benchmarking results suggest. Nonetheless, performance of the GPT models increases under both prompts, regardless of the evaluation metrics of FSA or FAA, suggesting that FAA could likely be a proxy for FSA.

For all GPT models, providing the C \& H list significantly helps with FSA, even when FAA stays at a similar level (among the respective problem subset). By comparison, PaLM 2 M could not benefit from additional information. These results imply the necessity for the finer-grained evaluations as we test LLMs on more challenging benchmarks.

\section{Experiment 2: Solution Verification}

The first wrong step (FWS) annotation allows for evaluating whether LLMs can read a solution to a problem and \textit{verify} its correctness, which is a much more difficult task than comparing the final {answer} against the ground-truth answer. In this set of experiments, we study how well LLMs can judge model-generated solutions, as well as the (error-free) ground truth solution.

\begin{figure}[!t]
    \centering

    \includegraphics[width=\columnwidth]{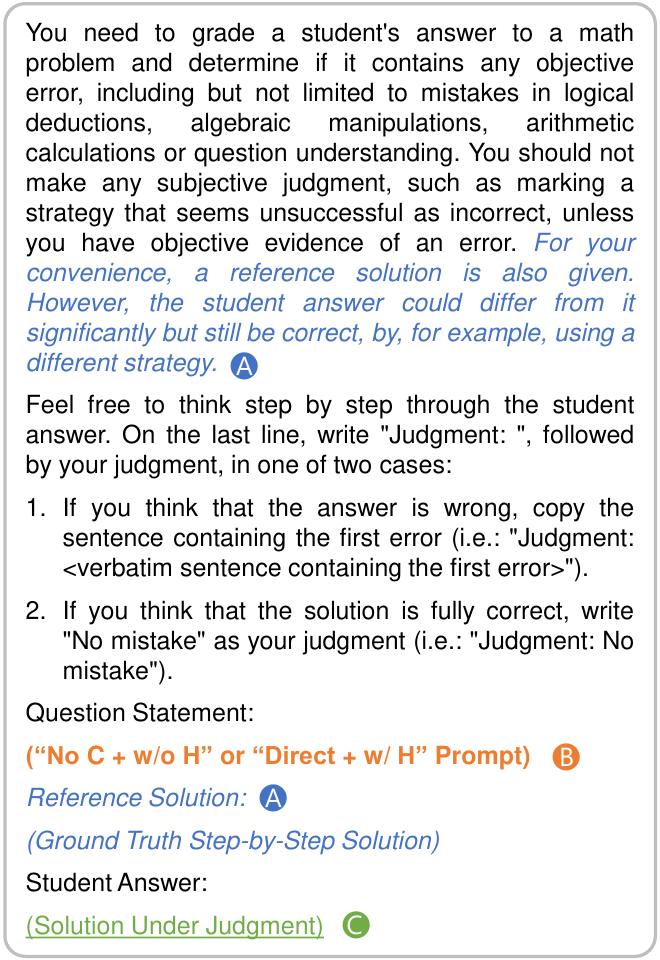}    
    \vspace{-5mm}
    
    \caption{The prompt for model verification evaluation. Text in black is given as the default, and we experiment with several variations. {\scriptsize \myCircled{A}}: we choose to give or withhold the reference solution in the prompt, where the \textit{\textcolor[RGB]{68, 114, 196}{blue italic texts}} are not provided in the latter case. {\scriptsize \myCircled{B}}: we evaluate model solutions for two prompts -- problem only and problem with concept and hint list. {\scriptsize \myCircled{C}}: the corresponding solution is given as the ``Student Answer''.}
\vspace{-6mm}
    
    \label{fig:verification-prompt}
\end{figure}

\subsection{Experimental Setup}
As noted in Sec.~\ref{sec:dataset}, the FWS dataset is obtained by (1) randomly assigning a model (out of GPT-3.5, GPT-4, GPT-4 Turbo and PaLM 2 M) to each problem and collecting its solutions for two prompts---Problem Only (i.e., ``No C + w/o H'' in Tab.~\ref{tab:result-table}) and Problem + C \& H (i.e., ``Direct + w/H'' in Tab.~\ref{tab:result-table})---and then (2) manually annotating the first wrong step  in each solution (or label it to be fully correct if there are no wrong steps). We first evaluate the ten models on each annotated solution, using the prompt shown in Fig.~\ref{fig:verification-prompt}. Two variants are explored: one where the reference solution is not given (by removing all \textcolor[RGB]{68, 114, 196}{\textit{blue italic texts}}) and the other where the reference solution is given (by using the complete prompt). This prompt both allows the model to engage in chain-of-thought reasoning and moreover enables easy parsing of the output, via the line starting with ``Judgment:''.

\subsection{Results}
\paragraph{Output judgment type.} We first study the judgment types from models (regardless of judgment correctness), with three categories: \textit{has-FWS}, where the model identifies a FWS, \textit{no-mistake}, where the model outputs ``Judgment: No mistake'', and \textit{invalid}, where the sentence following ``Judgment:'' is not in the solution or this line is not found. 

Tab.~\ref{tab:fws-response-type} lists the count breakdown for each of the three judgment types, in that respective order. Compared to the ground truth distribution, only GPT-4 and 4 Turbo behave reasonably, with other models either producing a large number of invalid judgments (e.g., Llama 2 7B and Mistral 7B), or failing to identify mistakes in most solutions (e.g., PaLM 2 Medium and Llama 2 70B). The results indicate that despite the ability to get correct final answers on challenging math problems, most models lack strong verification abilities or even present difficulty understanding and following the verification task prompt. For more detailed analyses, we focus on GPT-4 and 4 Turbo, which produce few invalid answers and best resemble the ground truth judgement patterns.

\begin{table}[!t]
    \centering
    \resizebox{\columnwidth}{!}{
    \begin{NiceTabular}{r|rr|rr}\toprule
        & \multicolumn{2}{c|}{Problem Only} & \multicolumn{2}{c}{Problem + C \& H}\\
        & w/o Ref Soln & w/ Ref Soln & w/o Ref Soln & w/ Ref Soln\\\midrule
        GPT-3.5 & 6 | 228 | \phantom{0}36 & 5 | 170 | \phantom{0}95 & 2 | 243 | \phantom{0}25 & 1 | 186 | \phantom{0}83 \\
        GPT-4 & 180 | \phantom{0}86 | \phantom{00}4 & 178 | \phantom{0}87 | \phantom{00}5 & 171 | \phantom{0}94 | \phantom{00}5 & 177 | \phantom{0}87 | \phantom{00}6 \\
        GPT-4 T & 203 | \phantom{0}56 | \phantom{0}11 & 200 | \phantom{0}55 | \phantom{0}15 & 205 | \phantom{0}49 | \phantom{0}16 & 193 | \phantom{0}67 | \phantom{0}10 \\
        PaLM 2 M & 0 | 233 | \phantom{0}37 & 0 | 225 | \phantom{0}45 & 0 | 221 | \phantom{0}49 & 0 | 236 | \phantom{0}34 \\
        Llama 2 7B & 124 | \phantom{0}20 | 126 & 106 | \phantom{0}54 | 110 & 149 | \phantom{0}11 | 110 & 130 | \phantom{0}27 | 113 \\
        Llama 2 70B & \phantom{00}0 | 264 | \phantom{00}6 & \phantom{00}0 | 255 | \phantom{0}15 & \phantom{00}0 | 264 | \phantom{00}6 & \phantom{00}0 | 266 | \phantom{00}4 \\
        Llama 3 8B & \phantom{00}3 | 252 | \phantom{0}15 & \phantom{0}13 | 249 | \phantom{00}8 & \phantom{00}3 | 252 | \phantom{0}15 & \phantom{0}18 | 241 | \phantom{0}11 \\
        Llama 3 70B & \phantom{0}93 | 128 | \phantom{0}49 & \phantom{0}46 | 199 | \phantom{0}25 & \phantom{0}69 | 138 | \phantom{0}63 & \phantom{0}47 | 197 | \phantom{0}26 \\
        Mistral 7B & \phantom{00}0 | \phantom{0}73 | 197 & \phantom{00}0 | \phantom{0}13 | 257 & \phantom{00}0 | \phantom{0}65 | 205 & \phantom{00}0 | \phantom{0}17 | 253 \\
        Mixtral 8x22B & \phantom{0}57 | 123 | \phantom{0}90 & \phantom{0}40 | 182 | \phantom{0}48 & \phantom{0}70 | 136 | \phantom{0}64 & \phantom{0}29 | 200 | \phantom{0}41 \\\midrule
        Ground truth & \multicolumn{2}{c}{238 | 32 | 0} & \multicolumn{2}{c}{206 | 64 | 0}\\\bottomrule
    \end{NiceTabular}
    }
    \vspace{-3mm}
    \caption{Count breakdown of judgment types produced by each model. In each cell, the three numbers represent the number of \textit{has-FWS}, \textit{no-mistake} and \textit{invalid} judgments. Each triplet sums up to 270, the total number of problems (and annotated solutions for the prompt). The ground truth  statistics are shown on the last line.}
\vspace{-2mm}
    
    \label{tab:fws-response-type}
\end{table}
\begin{table}[!t]
    \centering
    \resizebox{\columnwidth}{!}{
    \begin{NiceTabular}{lcc|p{1.1cm}@{\hspace{0.3ex}}p{1.1cm}@{\hspace{0.3ex}}p{1.1cm}@{\hspace{0.3ex}}p{1.1cm}|p{1.1cm}@{\hspace{0.3ex}}p{1.1cm}@{\hspace{-0.01cm}}p{0.01cm}}
    \CodeBefore
      \rectanglecolor{redbackground}{4-4}{11-4}
      \rectanglecolor{greenbackground}{4-5}{11-5}      
      \rectanglecolor{redbackground}{4-6}{11-8}      
      \rectanglecolor{greenbackground}{4-9}{11-10}
    \Body
    \toprule
        \multicolumn{3}{c|}{Setup} & \multicolumn{6}{c}{Model vs. Ground Truth Judgment}\\\midrule
        \rotatebox[x=10pt, y=10pt]{90}{Verifier} & \rotatebox[x=8pt, y=10pt]{90}{C \& H} & \rotatebox[x=13pt, y=10pt]{90}{Ref Soln} & \includegraphics[width=1.1cm]{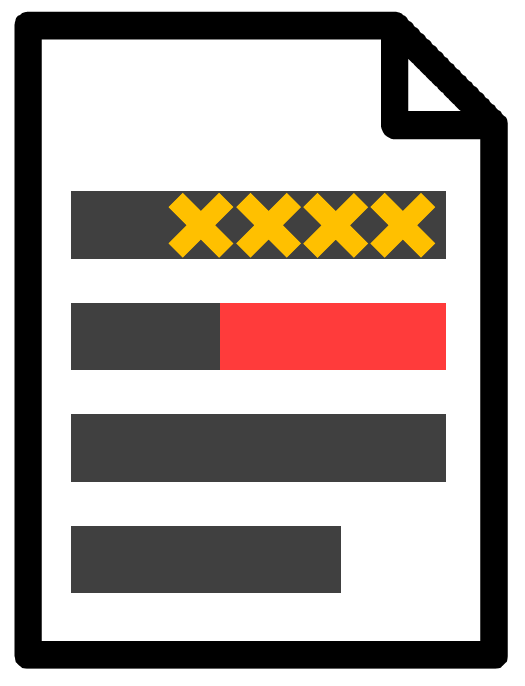} & \includegraphics[width=1.1cm]{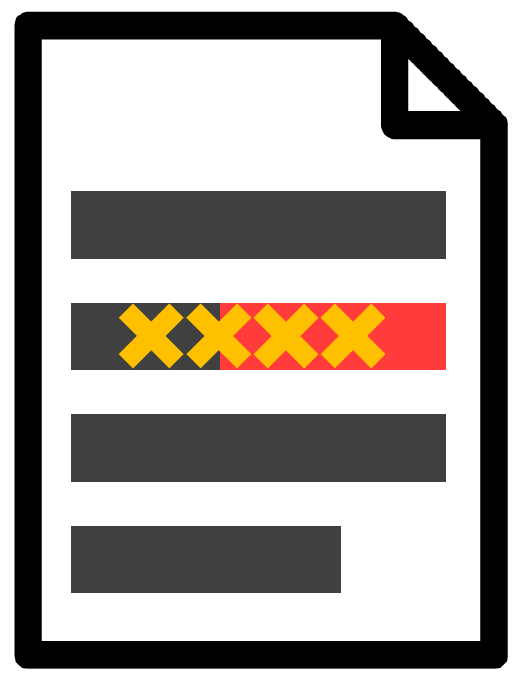} & \includegraphics[width=1.1cm]{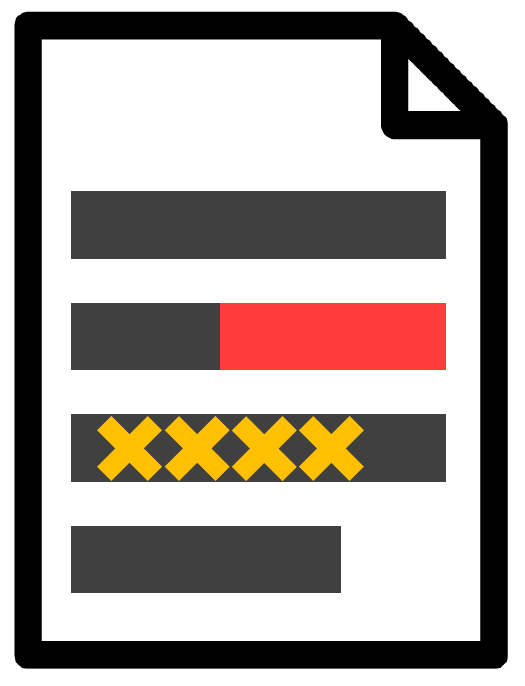}  & \includegraphics[width=1.1cm]{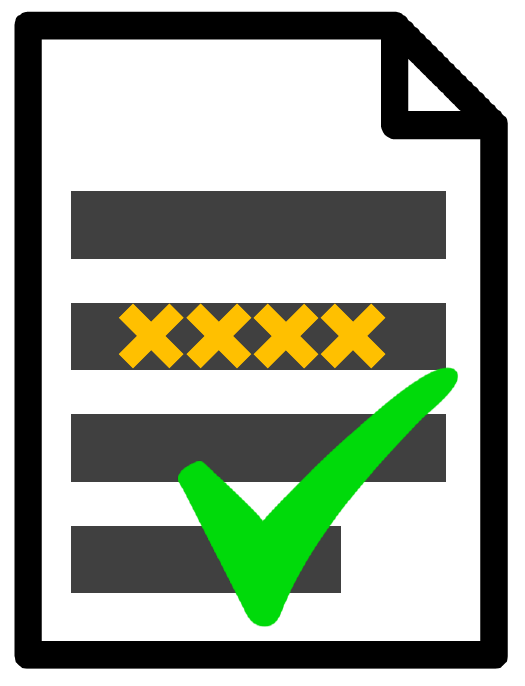} & \includegraphics[width=1.1cm]{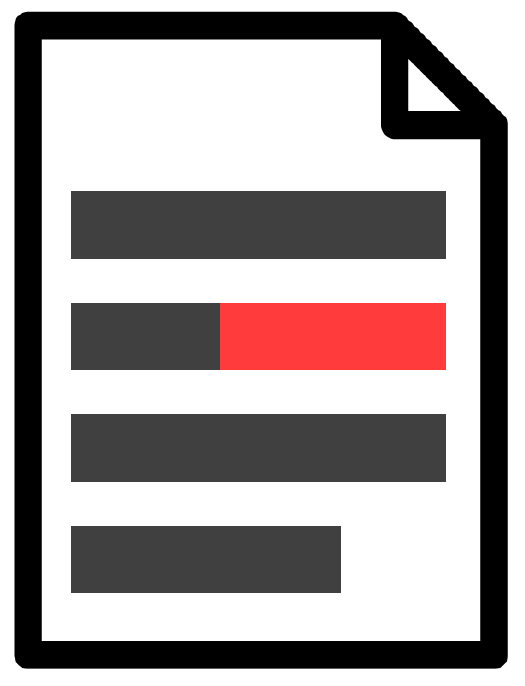} & \includegraphics[width=1.1cm]{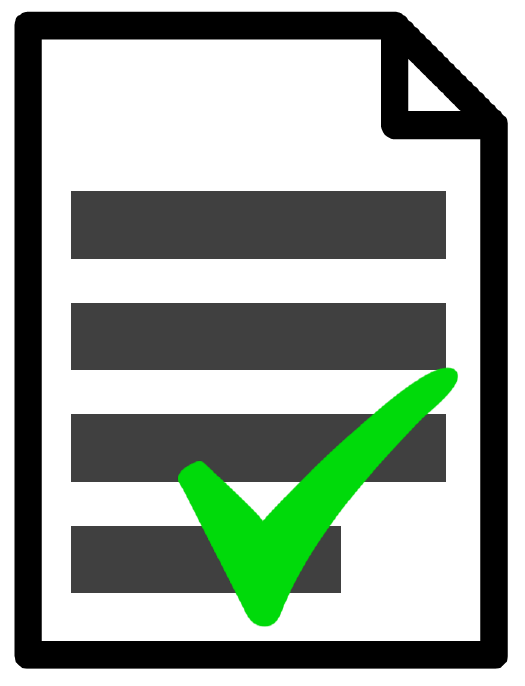}\\
        & & & \makecell[c]{early} & \makecell[c]{TP} & \makecell[c]{late} & \makecell[c]{spur.} & \makecell[c]{miss} & \makecell[c]{TN}\\\midrule
        \multirow{4}{*}{\rotatebox[origin=c]{90}{GPT-4}} & \xmark & \xmark & {\centering 33} & {\centering 61} & {\centering 76} & {\centering 10} & {\centering 64} & {\centering 22} & \\
        & \xmark & \cmark & {\centering 31} & {\centering 66} & {\centering 78} & {\centering \phantom{1}3} & {\centering 58} & {\centering 29} & \\
        & \cmark & \xmark & {\centering 33} & {\centering 50} & {\centering 69} & {\centering 19} & {\centering 49} & {\centering 45} & \\
        & \cmark & \cmark & {\centering 33} & {\centering 73} & {\centering 66} & {\centering \phantom{1}7} & {\centering 30} & {\centering 57} & \\\midrule
        \multirow{4}{*}{\rotatebox[origin=c]{90}{GPT-4 T}} & \xmark & \xmark & {\centering 33} & {\centering 71} & {\centering 86} & {\centering 13} & {\centering 37} & {\centering 19} & \\
        & \xmark & \cmark & {\centering 37} & {\centering 77} & {\centering 80} & {\centering 6} & {\centering 29} & {\centering 26} & \\
        & \cmark & \xmark & {\centering 31} & {\centering 54} & {\centering 87} & {\centering 33} & {\centering 19} & {\centering 30} & \\
        & \cmark & \cmark & {\centering 29} & {\centering 69} & {\centering 81} & {\centering 14} & {\centering 18} & {\centering 49} & \\\bottomrule
    \end{NiceTabular}
    }
    \caption{Detailed analysis of FWS identification. ``Verifier'' is the model under evaluation. A cross mark under ``C \& H'' uses the Problem Only (``No C + w/o H'') prompt and its solution, and a check mark uses the Problem + C \& H (``Direct + w/ H'') prompt. A cross mark under ``Ref Soln'' withholds the reference solution, and a check mark reveals it. For the six judgment types, red highlighting marks the ground truth (GT) FWS span, a green check mark means that the full solution is error-free, and yellow crosses mark the verifier's identification (VI), if present. The six judgments are: \textit{early} (where VI is before GT), \textit{TP} (true positive, where VI overlaps with GT), \textit{late} (where VI is after GP), \textit{spurious} (where GT is error-free but the verifier makes an identification), \textit{miss} (where verifier misses a GT FWS) and \textit{TN} (true negative, where model makes a correct judgment of error-free). \ctext{greenbackground}{Green} and \ctext{redbackground}{red} cell background colors indicates correct and incorrect judgments, respectively. The number in each cell counts the specific model judgments under the prompt. Invalid responses are excluded.}
    \label{tab:fws-position}
\end{table}

\paragraph{Judgment correctness analysis.}

For syntactically valid judgements (\textit{has-FWS} and \textit{no-mistake}), we evaluate their outputs in more depth. For each of the verifier model of GPT-4 and 4 Turbo, we consider 4 setups by varying two factors: the Problem Only or Problem + C \& H prompt, and with or without reference solution. Tab.~\ref{tab:fws-position} shows the counts of different model vs. ground truth judgments. 

We compute sensitivity and specificity to quantify the two verifiers' FWS identification accuracy. Recall that true-positive (TP) is the case where the model correctly identifies the ground truth FWS (with any overlapping) in a wrong solution, and true-negative (TN) is the case where the model correctly reports a no-mistake judgment for an error-free solution. {Sensitivity} is defined as the fraction of TP among all wrong solutions, and {specificity} as the fraction of TN among all error-free solutions. The former measures the verifiers' ability of accurately locating a FWS, while the latter measures that of recognizing correct solutions.

\begin{table}[t]
    \centering
    \resizebox{\columnwidth}{!}{
    \begin{NiceTabular}{cr|cc|cc}\toprule
         & & \multicolumn{2}{c|}{Sensitivity} & \multicolumn{2}{c}{Specificity}\\
         Prompt & Verifier & w/o Ref & w/ Ref & w/o Ref & w/ Ref \\\midrule
         \multirow{2}{*}{\makecell{Prob\\ Only}} & GPT-4 & 16.3 & \textit{19.9} & 71.0 & \textit{90.3} \\
         & GPT-4 T & \textbf{21.6} & \textbf{\textit{28.1}} & \textbf{58.1} & \textbf{\textit{80.6}} \\\midrule
         \multirow{2}{*}{\makecell{Prob + \\C \& H}} & GPT-4 & 18.1 & \textit{32.6} & 70.3 & \textit{89.1} \\
         & GPT-4 T & \textbf{28.1} & \textbf{\textit{33.8}} & \textbf{47.6} & \textbf{\textit{77.8}} \\\bottomrule
    \end{NiceTabular}
    }
    \caption{Sensitivity and specificity for different verifiers and experimental setups.}
    \label{tab:sesp}
\end{table}

Tab.~\ref{tab:sesp} shows these statistics. This table, and the raw counts in Tab.~\ref{tab:fws-position}, reveal two trends. First, giving the verifier access to reference solutions (i.e., italicizing) helps its performance, as evidenced by the increase in both sensitivity and specificity. Second, GPT-4 has lower sensitivity but higher specificity than GPT-4 Turbo (in bold), meaning that it is less capable of identifying FWSs, but also less prone to hallucinating errors in error-free solutions.

\begin{table}[!b]
    \centering
    \resizebox{\columnwidth}{!}{
    \begin{NiceTabular}{r|rrr}
    \CodeBefore
    \rectanglecolor{redbackground}{2-2}{11-2}
    \rectanglecolor{greenbackground}{2-3}{11-3}
    \rectanglecolor{redbackground}{2-4}{11-4}
    \Body
    \toprule
        Judgment & Has-FWS & No-Mistake & Invalid\\\midrule
        GPT-3.5 & 4 & 186 & 80 \\
        GPT-4 & 190 & 79 & 1 \\
        GPT-4 T & 161 & 102 & 7 \\
        PaLM 2 M & 0 & 187 & 83 \\
        Llama 2 7B & 76 & 123 & 71 \\
        Llama 2 70B & 2 & 235 & 15 \\
        Llama 3 8B & 30 & 222 & 18 \\
        Llama 3 70B & 38 & 223 & 9 \\
        Mistral 7B & 0 & 4 & 266 \\
        Mixtral 8x22B & 97 & 139 & 74 \\\bottomrule
    \end{NiceTabular}
    }
    \caption{Model verification of (error-free) reference solutions. Only ``No-Mistake'' judgments are correct.}
    \label{tab:fws-ref-solution}
\end{table}

\paragraph{Verification of reference solutions.}

Finally, we evaluate whether models can verify the reference ground-truth solution (pretending it to be the student answer). Here the correct response is always ``no-mistake''. The results are summarized in Tab.~\ref{tab:fws-ref-solution}.
 The best verifier models in the last experiment, GPT-4 and GPT-4 Turbo, make the most number of wrong ``has-FWS'' identifications. GPT-4 Turbo performs better than GPT-4 but still hallucinates a mistake on more than half solutions. Many models other than GPT-4 and GPT-4 Turbo ostensibly perform well on this task. However, this is due to their inability to identify \textit{any} FWS (c.f., Tab.~\ref{tab:fws-response-type}), as shown in the previous experiment. These results collectively indicate that solution verification is beyond the current capabilities of many LLMs.

\section{Discussion} 
This paper presents CHAMP, the Concept and Hint-Annotated Math Problems dataset, along with annotations of logical correctness of model-generated solutions to each problem. The unique construction of CHAMP enables previously under-explored studies regarding context-related reasoning abilities as well as verification abilities of LLMs.

We investigate the mathematical reasoning abilities of 10 proprietary and open-source LLMs of various sizes and release dates through CHAMP. Even the best models are far from perfect, and many models are unable to incorporate useful concepts and problem-specific hints. Even when the final answer accuracy is high, a closer inspection of the reasoning steps reveals that many models may be accidentally arriving at the correct answer. 

Furthermore, all models struggle at solution verification, indicating that these models often generate but do not understand their solutions, similar to some of the findings of recent work \cite{west2023generative,qiu2023phenomenal}.  These results advocate for a more fine-grained and multi-faceted investigation of LLMs' mathematical reasoning capabilities.

\section{Limitations}

While CHAMP reveals intriguing behaviors of different models, there are several limitations. For one, we relied on extensive manual annotation for high-quality labeling, and hence the dataset currently contains only 270 problems.  There is also the risk of dataset contamination; i.e., the models we evaluate may have been trained on the original versions of the problems from CHAMP. However, to mitigate this we rewrote the problems to fit the design of the dataset, minimizing the impact of memorization. Finally, while we rely on automatic evaluation for final answer accuracy to enable scalable evaluation, this may not be perfect. Our manual grading results suggest, though, that GPT-4's automatic grading has a high accuracy of 97\%. 


\section*{Acknowledgments}
We thank the reviewers for their suggestions. This study was supported in part by funds from MIT-IBM Watson AI.

\balance{}

\bibliography{custom}
\bibliographystyle{acl_natbib}

\appendix
\onecolumn

\section{Problem Collection and Annotation Considerations}
\label{app:category}

\subsection{Number Theory}
A notable feature of number theory problems is that most of them are proof problems. We manage to convert most of them into problems asking for an answer, with examples listed in Table \ref{tab:nt}. In addition, there are some questions which require non-trivial factorization. Since LLMs are often bad at arithmetics above 100, we provide them directly as hints, such as $1971=27\times 73$.

\begin{table}[!htb]
    \centering
    \resizebox{\columnwidth}{!}{
    \begin{NiceTabular}{p{10cm}p{10cm}}\toprule
        Before & After \\\midrule
        Prove that $n^4+4^n$ can never be a prime number for integer $n>1$. & For how many integers $n$ in $\{1, 2, ..., 99\}$ is $n^4+4^n$ a prime number? (Answer: 1)\\\midrule
        Prove that $x^2+y^2+z^2=2xyz$ has no positive integer solutions. & Find all positive integer solutions to the equation $x^2+y^2+z^2=2xyz$. (Answer: No positive integer solutions)\\\midrule
        Prove that $323 \mid 20^n+16^n-3^n-1$ for even $n$. & What are possible values of $20^n+16^n-3^n-1 \ \mathrm{mod}\  323$ for even $n$? (Answer: 0 is the only possible value)\\\bottomrule
    \end{NiceTabular}
    }
    \caption{Conversion of proof problems into those with check-able answers.}
    \label{tab:nt}
\end{table}

\subsection{Polynomial}
Some polynomial problems require factorization or root finding involving nontrivial arithmetics, similar to number theory problems. To reduce errors in this process, we provide the relevant arithmetic calculation as hints, such as $264=6\times 44$ when factoring $v^2-50v+264=(v-6)(v-44)$.

In addition, there are several polynomial division and remainder problems, for which we provide the concrete definition as a concept (although all models could easily retrieve and explain this definition with a straightforward query of ``What is polynomial division and remainder?''): 
\begin{quote}
    When a polynomial $f(x)$ is divided by a polynomial $g(x)$, the quotient $q(x)$ and the remainder $r(x)$ are polynomials such that $f(x)=g(x)q(x)+r(x)$ and the remainder $r(x)$ has degree less than that of $g(x)$.
\end{quote}

\subsection{Sequence}
A common type of problems in sequence is to find its limit. However, a prerequisite is to prove that the limit exists. Thus, we frame such questions explicitly, using wording such as ``Determine if the limit exists, and if so, find its value.'' We also annotate these questions with concepts stating the theorem that establish the existence of the limit, most commonly the monotone convergence theorem: 

\begin{quote}
    A sequence that is monotonic and bounded has a limit. Specifically, a sequence that is monotonically increasing and bounded from above, or monotonically decreasing and bounded from below, has a limit.
\end{quote}

In addition, a common strategy is induction, which shows that a property holds for all $a_n$ by showing that it holds for $a_n$ if it holds for all of $a_1, ..., a_{n-1}$. Because the instantiation of the strategy, especially the property to show, is problem-specific, we provide it as a hint, rather than a concept. 

\subsection{Inequality}
Just like with the category of number theory problems, many problems in inequality are written as proofs of inequality identity. We manage to convert them into questions requiring numerical answers with approaches such as asking for the extremum (i.e., maximum or minimum depending on the original inequality) value (while making sure that the value can indeed be attained by some variable value assignment). Some sample conversions are listed in Tab.~\ref{tab:ine}.

\begin{table*}[!htb]
    \centering
    \resizebox{\textwidth}{!}{
    \begin{tabular}{p{10cm}p{10cm}}\toprule
        Before & After \\\midrule
        Prove that, for $a, b, c>0$, $\sqrt[3]{abc} \leq \sqrt{(ab+bc+ca)/3}$? & 
        For positive $a, b, c$, what is the smallest value of $\sqrt{ab+bc+ac}/\sqrt[3]{abc}$? (Answer: $\sqrt{3}$)\\\midrule
        If $n>1$, proof that $1/(n+1)+1/(n+2)...+1/(2n)>1/2$. & For how many values of $n$ in $\{101, ..., 1000\}$ is $1/(n+1)+1/(n+2)+...+1/(2n)>1/2$? (Answer: 900)\\\midrule
        The product of three positive reals is 1. Their sum is greater than the sum of their reciprocals. Prove that exactly one of these numbers is $>1$. & The product of three positive real numbers is 1, and their sum is greater than the sum of their reciprocals. How many of them can be greater than 1? (Answer: 1)
        \\\bottomrule
    \end{tabular}
    }
    \caption{Conversion of inequality proof problems into those requiring answers.}
    \label{tab:ine}
\end{table*}

\subsection{Combinatorics}
Most combinatorics problems describe real-world scenarios. Where applicable, we provide any unmentioned commonsense knowledge (e.g., ``\textit{On a chess board, two rooks are placed peacefully if they are not on the same row or column.}'') before the problem (e.g., ``\textit{For an $n\times n$ chess board, find the number of ways that $n$ rooks can be placed peacefully (i.e., any two are placed peacefully), as an expression of n.}'').

In addition, many combinatorics problems ask for the number of ways in a setup size $n$ (e.g., the number of ways that $n$ horses can finish in a race with the possibility of ties), and it is solved in the following manner: 
\begin{enumerate}[parsep=0pt]
    \item Find a recurrence relationship to express $P(n)$ in terms of $P(n-1)$ and $P(n-2)$ (and possibly more terms), where $P(n)$ is the quantity asked in the question.
    \item Find the initial values $P(1), P(2)$ (and possibly more terms).
    \item Set up a characteristic equation (which is a polynomial) and find its root. 
    \item Use the roots to express $P(n)$ as a function of $n$. 
\end{enumerate}
The key difficulty is the root-finding part, so instead of asking for the general expression of $P(n)$ in terms of $n$, we ask for a specific value, such as $P(7)$, which could be worked out by repeatedly applying the recurrence relationship from the initial values. We also make sure that the asked $P(n)$ value is relatively small, usually less than 200, to minimize the chance of arithmetic errors. 

\clearpage

\section{Dataset Details}
\label{app:dataset}

Tab.~\ref{tab:sample-problem} shows one problem from each category, with problem, concepts and hints on the left column, and solution on the right column.

\begin{table}[!htb]
    \centering
    \resizebox{\textwidth}{!}{
    \begin{tabular}{p{12cm}|p{12cm}}\toprule\midrule
        \multicolumn{2}{c}{Number Theory (Problem ID: P\_Number-Theory\_1)}\\\midrule
         Problem: Are there integer solutions to the equation $(x^2-1)(y^2-1)+1985=z^2$? \rule[-10pt]{0pt}{0pt} \newline \textbf{C}oncepts and \textbf{H}ints: \begin{enumerate}[leftmargin=2em, topsep=0pt, itemsep=0pt, parsep=0pt]
         \item[H1.] Consider the equation modulo 9.
         \item[C1.] For integer $x$, $x^2 \bmod 9$ can take values of 0, 1, 4 and 7.
         \item[C2.] $(a+b) \bmod m=((a \bmod m)+(b \bmod m) \bmod m). (a-b) \bmod m=((a \bmod m)-(b \bmod m) \bmod m). ab \bmod m=((a \bmod m)(b \bmod m) \bmod m). a^k \bmod m=((a \bmod m)^k \bmod m)$.
         \item[H2.] $1985 \bmod 9=5$.
         \end{enumerate} & 
         Answer: No \rule[-10pt]{0pt}{0pt} \newline Solution Steps [Concepts and Hints Used]:
         \begin{enumerate}[leftmargin=*, topsep=0pt, itemsep=0pt, parsep=0pt]
             \item If the equation has a solution, we have $(x^2-1)(y^2-1)=z^2-1985$.
             \item Since $u^2 \bmod 9\in \{0, 1, 4, 7\}$, we have $u^2-1 \bmod 9\in \{0, 3, 6, 8\}$ and $(x^2-1)(y^2-1) \bmod 9\in \{0, 1, 3, 6\}$. [H1, C1, C2]
             \item However, since $1985 \bmod 9=5$, $z^2-1985 \bmod 9\in \{2, 4, 5, 8\}$. [H1, H2, C2]
             \item Since there is no overlapping values, we conclude that the equation has no solution.
         \end{enumerate}
        \\\midrule\midrule
         \multicolumn{2}{c}{Polynomial (Problem ID: P\_Polynomial\_1)}\\\midrule
         Problem: What is the remainder of $nx^{n+1}-(n+1)x^n+1$ divided by $(x-1)^2$?\rule[-10pt]{0pt}{0pt} \newline \textbf{C}oncepts and \textbf{H}ints: \begin{enumerate}[leftmargin=2em, topsep=0pt, itemsep=0pt, parsep=0pt]
        	\item[C1.] When a polynomial $f(x)$ is divided by a polynomial $g(x)$, the quotient $q(x)$ and the remainder $r(x)$ are polynomials such that $f(x)=g(x)q(x)+r(x)$ and the remainder $r(x)$ has degree less than that of $g(x)$.
         \item[H1.] Let $f(x)=nx^{n+1}-(n+1)x^n+1$ and study $f(1)$ and $f'(1)$. \end{enumerate} & 
         Answer: 0\rule[-10pt]{0pt}{0pt} \newline Solution Steps [Concepts and Hints Used]:
         \begin{enumerate}[leftmargin=*, topsep=0pt, itemsep=0pt, parsep=0pt]
             \item We have $f(x)=(x-1)^2*q(x)+r(x)$ where $r(x)$ is a polynomial of degree at most 1 (i.e., $r(x)=ax+b$). [C1]
             \item Thus, $f(1)=0=r(1)$. [H1]
             \item We have $f'(x)=2(x-1)*q(x)+(x-1)^2*q(x)+r'(x), so f'(1)=r'(1)=0$. [H1]
             \item Since $r(x)$ has the form of $ax+b$, we have $a+b=0$, $a=0$, so $b=0$.
             \item Thus, $r(x)=0$ is the remainder.
         \end{enumerate}
         \\\midrule\midrule
         \multicolumn{2}{c}{Sequence (Problem ID: P\_Sequence\_2)}\\\midrule
         Problem: Let $\{x_n\}, \{y_n\}, \{z_n\}$ be three sequences with positive initial terms $x_1, y_1, z_1$, defined as $x_{n+1}=y_n+1/z_n, y_{n+1}=z_n+1/x_n, z_{n+1}=x_n+1/y_n$. Let $w_n$ be the maximum value of $x_n, y_n, z_n$. For different values of $x_1, y_1, z_1$, do we have $w_{200}$ always greater than 20, always smaller than 20, or sometimes greater and sometimes smaller than 20? \rule[-10pt]{0pt}{0pt} \newline \textbf{C}oncepts and \textbf{H}ints: \begin{enumerate}[leftmargin=2em, topsep=0pt, itemsep=0pt, parsep=0pt]
        \item[H1.] Let $a_n=x_n+y_n+z_n$.
        \item[H2.] Derive a lower bound on $a_2$.
        \item[C1.] For positive $x$, $x+1/x\geq 2$, with equality if and only if $x=1$.
        \item[H3.] Compare $a_n$ with $18n$ for all $n$.
        \item[C2.] $(x\pm y)^2=x^2\pm 2xy+y^2$.
        \item[C3.] For real numbers $a_1, ..., a_n$ and $b_1, ..., b_n$, $(a_1b_1+...+a_nb_n)^2\leq (a_1^2+...+a_n^2)(b_1^2+...+b_n^2)$.
         \end{enumerate} & 
         Answer: Always greater than 20 \rule[-10pt]{0pt}{0pt} \newline Solution Steps [Concepts and Hints Used]:
         \begin{enumerate}[leftmargin=*, topsep=0pt, itemsep=0pt, parsep=0pt]
             \item 	Let $a_n=x_n+y_n+z_n$. [H1]
             \item We have $a_2^2=(x_1+1/x_1+y_1+1/y_1+z_1+1/z_1)^2\geq (2+2+2)^2=36=2\cdot 18$. [H2, C1]
             \item If $a_n^2\geq 18n$, then we have $a_{n+1}^2=(x_n+1/x_n+y_n+1/y_n+z_n+1/z_n)^2\geq a_n^2+2(x_n+y_n+z_n)(1/x_n+1/y_n+1/z_n)\geq a_n^2+2*9\geq 18n+18=18(n+1)$. [H3, C2, C3]
             \item So we have $a_n^2\geq 18n$. [H3]
             \item Thus, $a_{200}^2\geq 18\cdot 200=3600$, which means that $a_{200}=x_{200}+y_{200}+z_{200}\geq 60$.
             \item So one of $x_{200}, y_{200}, z_{200}$ must be at least 20.
             \item Thus, $w_{200}$ cannot be smaller than 20.
         \end{enumerate}\\\midrule\midrule
         \multicolumn{2}{c}{Inequality (Problem ID: P\_Inequality\_2)}\\\midrule
         Problem: For positive $a, b$, what is the smallest value of $(a^2+b^2)/(a+b)^2$? \rule[-10pt]{0pt}{0pt} \newline \textbf{C}oncepts and \textbf{H}ints: \begin{enumerate}[leftmargin=2em, topsep=0pt, itemsep=0pt, parsep=0pt]
        	\item[C1.] For non-negative $x, y$, we have $(x+y)/2\leq \sqrt{(x^2+y^2)/2}$, with equality if and only if $x=y$.
         \end{enumerate} & 
         Answer: $1/2$ \rule[-10pt]{0pt}{0pt} \newline Solution Steps [Concepts and Hints Used]:
         \begin{enumerate}[leftmargin=*, topsep=0pt, itemsep=0pt, parsep=0pt]
             \item Since $(a+b)/2\leq \sqrt{(a^2+b^2)/2}$, we have $(a+b)^2/4\leq (a^2+b^2)/2$. [C1]
             \item This means that $(a^2+b^2)/(a+b)^2\geq 1/2$.
             \item So the smallest value is $1/2$, achieved at $a=b$.
         \end{enumerate}
         \\\midrule\midrule
         \multicolumn{2}{c}{Combinatorics (Problem ID: P\_Combinatorics\_1)}\\\midrule
        Problem: Let a string consist of digit 1, 2, 3. How many such strings of length 6 have adjacent digit differing by less than or equal to 1?\rule[-10pt]{0pt}{0pt} \newline \textbf{C}oncepts and \textbf{H}ints: \begin{enumerate}[leftmargin=2em, topsep=0pt, itemsep=0pt, parsep=0pt]
        	\item[H1.] Let $x_n$, $y_n$, $z_n$ be the number of length-$n$ strings that end with digit 1, 2, 3 respectively.
            \item[H2.] What are $x_1$, $y_1$, $z_1$?
            \item[H3.] By appending a digit to the existing string, derive the formula for $x_{n+1}$, $y_{n+1}$, $z_{n+1}$ from $x_n$, $y_n$, $z_n$.
            \item[C1.] If there are n actions, with $p_i$ ways to perform the $i$-th action, and no two actions can be performed at the same time, then there are $p_1+p_2+...+p_n$ ways to perform the action in total. 
        \end{enumerate}& Answer: 239\rule[-10pt]{0pt}{0pt} \newline Solution Steps [Concepts and Hints Used]:
        \begin{enumerate}[leftmargin=*, topsep=0pt, itemsep=0pt, parsep=0pt]
            \item Let $x_n$, $y_n$, $z_n$ be the number of length-$n$ strings that end with digit 1, 2, 3 respectively. [H1]
            \item Thus, we have $x_1=y_1=z_1=1$. [H2]
            \item For a string ending with 1, we can append 1 and 2; for a string ending with 2, we can append 1, 2 and 3; for a string ending with 3, we can append 2 and 3. [H3]
            \item Thus, we have $x_{n+1}=x_n+y_n, y_{n+1}=x_n+y_n+z_n, and z_{n+1}=y_n+z_n$. [H3, C1]
            \item Starting from (1, 1, 1), we have the sequence of $(x_n, y_n, z_n)$ to be (1, 1, 1), (2, 3, 2), (5, 7, 5), (12, 17, 12), (29, 41, 29), (70, 99, 70).
            \item Thus, in total, there are $x_6+y_6+z_6=70+99+70=239$ such strings.
        \end{enumerate}
         \\\bottomrule
    \end{tabular}
    }
    \caption{One example problem per category from the dataset. The left column presents the problem, along with relevant concepts and hints. The right column gives the solution, both the final answer and the full step-wise solution with concept and hint labels. }
    \label{tab:sample-problem}
\end{table}

\FloatBarrier

\clearpage

Table \ref{tab:baseline} presents a breakdown of problems by answer formats and the corresponding baseline answer design. This baseline simulates the best performance of a dummy model that has no math reasoning ability but can answer the question in a semantically sensible manner (e.g., answering Yes or No to a question asking for a Boolean answer). 

\begin{table}[!htb]
    \centering
    \resizebox{\columnwidth}{!}{
    \begin{NiceTabular}{llp{9cm}p{4cm}}\toprule
        Answer Format & \# Probs & Example & Baseline Answer \\\midrule
        Boolean & 42 & Is $4^{545}+545^4$ a prime number? & No\\\midrule
        Numeric & 162 & In how many ways can 4 horses go through the finish (with possibility of ties)? & 0\\\midrule
        Expression & 45 & 	Among all sequences of positive integer numbers have sum $n$, for integer $k<n-1$, how many times does the number $k$ appear, as an expression of $n$ and $k$? & Sum of all variables (i.e., $n+k$ for the example)\\\midrule
        Enumeration & 21 & Find all integer solutions to the equation $15x^2-7y^2=9$. & None\\\bottomrule
    \end{NiceTabular}
    }
    \caption{The construction of baseline answers based on four answer formats.}
    \label{tab:baseline}
\end{table}

\section{Full Prompt Texts}
\label{app:prompts}

This section lists all the prompts used in the experiments. Texts of normal fonts are provided literally. Parenthesized italic \textit{(texts)} are provided to the model, and parenthesized bold \textbf{(texts)} are model outputs.

Tab.~\ref{tab:prompt-problem-solving} shows the main prompt setup for the model's problem solving capability evaluation. The final answer summary is evaluated by GPT-4 with the prompt listed in Tab.~\ref{tab:prompt-grading}. We include ``Partially correct'' as one grading verdict for the model to use in ambiguous situations (e.g., the solver model finds one of two solutions to the equation) but treat it as incorrect for accuracy calculation (e.g., in Tab.~\ref{tab:result-table}).

\begin{table}[!htb]
    \centering
    \resizebox{\textwidth}{!}{
    \begin{NiceTabular}{>{\raggedleft\arraybackslash}p{1.2cm}p{18cm}}
    \CodeBefore
      \rectanglecolor{orange}{3-1}{4-2}
    \Body\toprule
        Role & Message \\\midrule
        System & You are an expert on mathematics.\\\midrule
        User & \multirow{2}{*}{\parbox{18cm}{\textit{\textbf{(One or more rounds of user-solver conversation that end in the solver generating the full solution as the message of the last round. The specific conversation contents are presented in Tab.~\ref{tab:prompt-0shot}-\ref{tab:prompt-problemC}.)}}}}\\
        Solver \\\midrule
        User & Now, summarize the answer above in one sentence, without any intermediate steps or explanations. \\\midrule
        Solver & \textbf{(Model-generated final answer summary)} \\\bottomrule
    \end{NiceTabular}
    }
    \caption{Prompt for eliciting full solution and final answer summary from the model under evaluation (i.e., solver).}
    \label{tab:prompt-problem-solving}
\end{table}

\begin{table}[!htb]
    \resizebox{\textwidth}{!}{
    \begin{NiceTabular}{>{\raggedleft\arraybackslash}p{1.2cm}p{18cm}}\toprule
        Role & Message \\\midrule
        System & You are a math teacher and need to grade student's homework. For each question, you need to determine the correctness of the student answer, given the reference answer. You only need to judge the correctness of the final answer, and should not consider any reasoning or explanations given in the answer. Note that if the student gives an obviously equivalent solution to the reference answer (e.g., $1.5$ vs $3/2$ or $a^2-b^2$ vs $(a+b)(a-b)$), the answer should be judged as correct. Your decision should be one of ``Correct'', ``Incorrect'' or ``Partially correct''. There is no need to explain your decision. \\\midrule
        User & The question is:\newline\textit{(Problem statement)}\newline The reference answer is:\newline\textit{(Ground truth final answer from the dataset)}\newline The student answer is:\newline\textit{(Model-generated final answer summary)}\newline Is the student answer correct, incorrect, or partially correct?\\\midrule
        GPT-4 & \textbf{(Grading verdict)}\\\bottomrule
    \end{NiceTabular}
    }
    \caption{Prompt for grading the solver's final answer summary using GPT-4.}
    \label{tab:prompt-grading}
\end{table}

\FloatBarrier
\clearpage

Tab.~\ref{tab:prompt-0shot}-\ref{tab:prompt-partial} present the prompts for evaluating models under current practices, including zero-shot, few-shot (5-shot) and zero-shot with partial solution. These conversations are to be swapped into the \ctext{orange}{orange cell} of Tab.~\ref{tab:prompt-problem-solving}. Note that in the few-shot prompt of tab.~\ref{tab:prompt-5shot}, only the last round is actually generated by the model. The ``solver output'' in the earlier rounds are directly fed to the model as the context (pretending to be earlier model generations).

\begin{table}[!htb]
    \resizebox{\textwidth}{!}{
    \begin{tabular}{>{\raggedleft\arraybackslash}p{1.2cm}p{18cm}}\toprule
        Role & Message \\\midrule
        User & Solve the following problem. Make sure to show your work before giving the final answer.\newline\textit{(Problem statement)}\\\midrule
        Solver & \textbf{(Model-generated full solution)}\\\bottomrule
    \end{tabular}
    }
    \caption{Zero-shot prompt.}
    \label{tab:prompt-0shot}
\end{table}

\begin{table}[!htb]
    \resizebox{\textwidth}{!}{
    \begin{tabular}{>{\raggedleft\arraybackslash}p{1.2cm}p{18cm}}\toprule
        Role & Message \\\midrule
        User & Solve the following problem. Make sure to show your work before giving the final answer.\newline\textit{(Statement of sample problem 1)}\\\midrule
        Solver & \textit{(Ground truth solution steps for problem 1)}\\\midrule
        User & Solve the following problem. Make sure to show your work before giving the final answer.\newline\textit{(Statement of sample problem 2)}\\\midrule
        Solver & \textit{(Ground truth solution steps for problem 2)}\\\midrule
        User & Solve the following problem. Make sure to show your work before giving the final answer.\newline\textit{(Statement of sample problem 3)}\\\midrule
        Solver & \textit{(Ground truth solution steps for problem 3)}\\\midrule
        User & Solve the following problem. Make sure to show your work before giving the final answer.\newline\textit{(Statement of sample problem 4)}\\\midrule
        Solver & \textit{(Ground truth solution steps for problem 4)}\\\midrule
        User & Solve the following problem. Make sure to show your work before giving the final answer.\newline\textit{(Statement of sample problem 5)}\\\midrule
        Solver & \textit{(Ground truth solution steps for problem 5)}\\\midrule
        User & Solve the following problem. Make sure to show your work before giving the final answer.\newline\textit{(Problem statement)}\\\midrule
        Solver & \textbf{(Model-generated full solution)}\\\bottomrule
    \end{tabular}
    }
    \caption{Few-shot prompt.}
    \label{tab:prompt-5shot}
\end{table}

\begin{table}[!htb]
    \resizebox{\textwidth}{!}{
    \begin{tabular}{>{\raggedleft\arraybackslash}p{1.2cm}p{18cm}}\toprule
        Role & Message \\\midrule
        User & Solve the following problem. Make sure to show your work before giving the final answer.\newline\textit{(Problem statement)}\newline Below is a partial solution to the problem that may be helpful:\newline \textit{(List of steps revealed in the partial solution)}\\\midrule
        Solver & \textbf{(Model-generated full solution)}\\\bottomrule
    \end{tabular}
    }
    \caption{Zero-shot prompt with partial solution provided.}
    \label{tab:prompt-partial}
\end{table}

\FloatBarrier
\clearpage

Tab.~\ref{tab:prompt-noC}-\ref{tab:prompt-problemC} presents the prompts for different concept provision methods covered in Tab.~\ref{tab:result-table}. The \ctext{lightgreen}{light green texts} and \ctext{darkgreen}{dark green texts} are used for the \ctext{lightgreen}{w/o H} and \ctext{darkgreen}{w/ H} prompts respectively, consistent with the color-coding of Tab.~\ref{tab:result-table}. Any corner cases are discussed in table captions.

\begin{table}[!htb]
    \resizebox{\textwidth}{!}{
    \begin{tabular}{>{\raggedleft\arraybackslash}p{1.2cm}p{18cm}}\toprule
        Role & Message \\\midrule
        User & Solve the following problem. Make sure to show your work before giving the final answer.\newline\textit{(Problem statement)}\newline \newline \ctext{darkgreen}{You may find the following information useful:}\newline
        \ctext{darkgreen}{\textit{(List of all hints.)}}\\\midrule
        Solver & \textbf{(Model-generated full solution)}\\\bottomrule
    \end{tabular}
    }
    \caption{The ``No C'' concept provision prompt (i.e. not providing any concept). The \ctext{lightgreen}{w/o H} version (and \ctext{darkgreen}{w H} version when the problem does not have any hint) is the same as the zero-shot prompt of Tab.~\ref{tab:prompt-0shot}. }
    \label{tab:prompt-noC}
\end{table}

\begin{table}[!htb]
    \resizebox{\textwidth}{!}{
    \begin{tabular}{>{\raggedleft\arraybackslash}p{1.2cm}p{18cm}}\toprule
        Role & Message \\\midrule
        User & Solve the following problem. Make sure to show your work before giving the final answer.\newline\textit{(Problem statement)}\newline \newline You may find the following information useful:\newline
        \ctext{lightgreen}{\textit{(List of relevant concepts.)}} / 
        \ctext{darkgreen}{\textit{(List of relevant concepts and hints.)}}\\\midrule
        Solver & \textbf{(Model-generated full solution)}\\\bottomrule
    \end{tabular}
    }
    \caption{The ``Direct'' concept provision prompt (also used for ``Root'' and ``Misleading'' with respective concepts). If there are no concepts in the \ctext{lightgreen}{w/o H} version, then the last paragraph is removed entirely and the prompt reduces to the zero-shot prompt of Tab.~\ref{tab:prompt-0shot}. }
    \label{tab:prompt-directC}
\end{table}

\begin{table}[!htb]
    \resizebox{\textwidth}{!}{
    \begin{tabular}{>{\raggedleft\arraybackslash}p{1.2cm}p{18cm}}\toprule
        Role & Message \\\midrule
        User & Please explain the following concept: \textit{(name of concept 1, skip this round if unnamed)}.\\\midrule
        Solver & \textbf{(Model-generated concept explanation)}\\\midrule
        User & Please explain the following concept: \textit{(name of concept 2, skip this round if unnamed)}.\\\midrule
        Solver & \textbf{(Model-generated concept explanation)}\\\midrule
        \multicolumn{2}{c}{\textbf{\textit{(One round of conversation for each named concept)}}}\\\midrule
        User & Please explain the following concept: \textit{(name of concept $n$, skip this round if unnamed)}.\\\midrule
        Solver & \textbf{(Model-generated concept explanation)}\\\midrule
        User & Solve the following problem. Make sure to show your work before giving the final answer.\newline\textit{(Problem statement)}\newline \newline Besides the concepts above, you may also find the following information useful:\newline
        \ctext{lightgreen}{\textit{(List of remaining unnamed concepts.)}} / 
        \ctext{darkgreen}{\textit{(List of remaining unnamed concepts and all hints.)}}\\\midrule
        Solver & \textbf{(Model-generated full solution)}\\\bottomrule
    \end{tabular}
    }
    \caption{The ``Name'' concept provision prompt. If there are no unnamed concepts (and hints), i.e., an empty list, then the sentence ``Besides the concepts above...'' is replaced with ``You may find the above concepts helpful.'', and the prompt is terminated.}
    \label{tab:prompt-nameC}
\end{table}

\begin{table}[!htb]
    \resizebox{\textwidth}{!}{
    \begin{tabular}{>{\raggedleft\arraybackslash}p{1.2cm}p{18cm}}\toprule
        Role & Message \\\midrule
        User & Please give an example that applies the following concept: \newline \textit{(Text of concept 1)}.\\\midrule
        Solver & \textbf{(Model-generated example)}\\\midrule
        User & Please give an example that applies the following concept: \newline \textit{(Text of concept 2)}.\\\midrule
        Solver & \textbf{(Model-generated example)}\\\midrule
        \multicolumn{2}{c}{\textbf{\textit{(One round of conversation for each concept)}}}\\\midrule
        User & Please give an example that applies the following concept: \newline \textit{(Text of concept $n$)}.\\\midrule
        Solver & \textbf{(Model-generated concept explanation)}\\\midrule
        User & Solve the following problem. Make sure to show your work before giving the final answer.\newline\textit{(Problem statement)}\newline 
        \newline \ctext{lightgreen}{You may find the above concepts helpful.}
        \newline \ctext{darkgreen}{Besides the concepts above, you may also find the following information useful:}\newline
        \ctext{darkgreen}{\textit{(List of all hints.)}}\\\midrule
        Solver & \textbf{(Model-generated full solution)}\\\bottomrule
    \end{tabular}
    }
    \caption{The ``Example'' concept provision prompt. If there are no hints, then we use the \ctext{lightgreen}{w/o H} version.}
    \label{tab:prompt-exampleC}
\end{table}

\begin{table}[!htb]
    \resizebox{\textwidth}{!}{
    \begin{tabular}{>{\raggedleft\arraybackslash}p{1.2cm}p{18cm}}\toprule
        Role & Message \\\midrule
        User & First, study the following problems and their solutions as they may contain helpful information.\newline
        \textit{(Sample problem that uses concept 1 and its step-wise solution)}
        \newline 
        \textit{(Sample problem that uses concept 2 and its step-wise solution)}
        \newline 
        \textit{(...)}
        \newline 
        \textit{(Sample problem that uses concept $n$ and its step-wise solution)}
        \newline\newline
        With what you have learned, solve the following problem. Make sure to show your work before giving the final answer.\newline\textit{(Problem statement)}\newline \newline \ctext{darkgreen}{In addition, you may also find the following information helpful:}\newline
        \ctext{darkgreen}{\textit{(List of all hints.)}}\\\midrule
        Solver & \textbf{(Model-generated full solution)}\\\bottomrule
    \end{tabular}
    }
    \caption{The ``Problem'' concept provision prompt. If there are no hints, then the \ctext{darkgreen}{colored text} is omitted. }
    \label{tab:prompt-problemC}
\end{table}

\FloatBarrier
\phantom{i}

\clearpage

\section{Model-Generated Solutions and First Wrong Step Annotations}
\label{app:fws-solution}

Tab.~\ref{tab:full-solution-1} and \ref{tab:full-solution-2} show one solution generated by each model. The final answer (FA) correctness and full solution (FS) correctness are separately noted. When the FS is incorrect (regardless of the FA correctness), the first wrong step (FWS) is highlighted in \textcolor{red}{red}, which consists the FWS corpus annotation. 

\begin{table}[!htb]
    \centering
    \resizebox{\textwidth}{!}{
    \begin{tabular}{p{5.7cm}|p{5.7cm}|p{5.7cm}}\toprule\midrule
Solution Author: GPT-3.5 & Final Answer: \textcolor{Green}{Correct} & Full Solution: \textcolor{red}{Incorrect} \\\midrule
\multicolumn{3}{l}{\parbox{18cm}{Problem ID: P\_Inequality\_10\rule[2ex]{0pt}{2pt}\newline Problem: For $x, y, z>0$, what is the smallest value of $x^2/y^2+y^2/z^2+z^2/x^2-(y/x+z/y+x/z)$?\newline  Answer: 0\rule[-1ex]{0pt}{2pt}}}\\\midrule
\multicolumn{3}{l}{\parbox{18cm}{
Model-generated solution:\rule[2ex]{0pt}{2pt}\newline 
To find the smallest value of the expression, we can use the AM-GM inequality.
\newline 
\textcolor{red}{By AM-GM, we have:}
\newline 
\textcolor{red}{$x^2/y^2 + y^2/z^2 + z^2/x^2 \geq 3\sqrt{(x^2/y^2)(y^2/z^2)(z^2/x^2)}$}
\newline 
Simplifying the right side, we get:
\newline 
$3\sqrt{(x^2/y^2)(y^2/z^2)(z^2/x^2)} = 3\sqrt{(x^2y^2z^2)/(x^2y^2z^2)} = 3\sqrt1 = 3$
\newline 
Now, let's consider the second term, $-(y/x + z/y + x/z)$.
\newline 
By AM-GM, we have:
\newline 
$y/x + z/y + x/z \geq 3\sqrt{(y/x)(z/y)(x/z)}$
\newline 
Simplifying the right side, we get:
\newline 
$3\sqrt{(y/x)(z/y)(x/z)} = 3\sqrt{(xyz)/(xyz)} = 3\sqrt 1 = 3$
\newline 
Therefore, the smallest value of the expression is:
\newline 
$3 - 3 = 0$
\newline 
So, the smallest value of $x^2/y^2 + y^2/z^2 + z^2/x^2 - (y/x + z/y + x/z)$ is 0.\rule[-1ex]{0pt}{2pt}
        }}\\\midrule\midrule
Solution Author: GPT-4 & Final Answer: \textcolor{Green}{Correct} & Full Solution: \textcolor{red}{Incorrect} \\\midrule
        \multicolumn{3}{l}{\parbox{18cm}{Problem ID: P\_Number-Theory\_13\rule[2ex]{0pt}{2pt}\newline Problem: What is the smallest value of $|36^m-5^n|$ for positive integers $m$ and $n$?\newline  Answer: 11\rule[-1ex]{0pt}{2pt}}}\\\midrule
        \multicolumn{3}{l}{\parbox{18cm}{
        Model-generated solution:\rule[2ex]{0pt}{2pt}\newline
To find the smallest value of $|36^m - 5^n|$ for positive integers $m$ and $n$, we need to find the smallest values of $m$ and $n$ such that $36^m$ is as close as possible to $5^n$.
\newline
Let's start with $m = 1$ and $n = 1$. We have:
\newline
$|36^1 - 5^1| = |36 - 5| = 31$
\newline
If we increase $m$ to 2, we get:
\newline
$|36^2 - 5^1| = |1296 - 5| = 1291$
\newline
This is much larger, so increasing $m$ is not a good strategy. Let's try increasing $n$ to 2:
\newline
$|36^1 - 5^2| = |36 - 25| = 11$
\newline
This is smaller, so increasing $n$ seems to be a good strategy. Let's try $n = 3$:
\newline
$|36^1 - 5^3| = |36 - 125| = 89$
\newline
This is larger again, so $n = 2$ seems to be the best choice. Let's check $m = 2$ just to be sure:
\newline
$|36^2 - 5^2| = |1296 - 25| = 1271$
\newline
\textcolor{red}{This is much larger, so $m = 1$ and $n = 2$ is the best choice.}
\newline
So, the smallest value of $|36^m - 5^n|$ for positive integers $m$ and $n$ is 11.\rule[-1ex]{0pt}{2pt}
        }}\\
        \bottomrule
    \end{tabular}
    }
    \caption{Sample solutions generated by different models and their correctness. If the full solution is not correct, the first wrong step (FWS) is marked in \textcolor{red}{red}. Table continued in Tab.~\ref{tab:full-solution-2}.}
    \label{tab:full-solution-1}
\end{table}

\begin{table}[!htb]
    \centering
    \resizebox{\textwidth}{!}{
    \begin{tabular}{p{5.7cm}|p{5.7cm}|p{5.7cm}}\toprule\midrule
Solution Author: GPT-4 T & Final Answer: \textcolor{Green}{Correct} & Full Solution: \textcolor{Green}{Correct} \\\midrule
\multicolumn{3}{l}{\parbox{18cm}{Problem ID: P\_Combinatorics\_40\rule[2ex]{0pt}{2pt}\newline Problem: Find the number of ways to fill a 2 x 11 rectangle with 2 x 1 tiles. \newline  Answer: 144\rule[-1ex]{0pt}{2pt}}}\\\midrule
\multicolumn{3}{l}{\parbox{18cm}{
Model-generated solution:\rule[2ex]{0pt}{2pt}\newline
To solve this problem, we can use a recursive approach. Let's denote the number of ways to fill a 2 x $n$ rectangle with 2 x 1 tiles as $f(n)$. We are looking to find $f(11)$.
\newline
Let's consider the first column of the 2 x 11 rectangle. We have two choices for placing the tiles:
\newline
1. We place two tiles\newline vertically in the first column. If we do this, we are left with a 2 x (11-1) = 2 x 10 rectangle to fill. The number of ways to fill the rest of the rectangle is $f(10)$.
\newline
2. We place one tile horizontally at the top of the first column. If we do this, we must also place another tile horizontally below it to fill the first two columns. This leaves us with a 2 x (11-2) = 2 x 9 rectangle to fill. The number of ways to fill the rest of the rectangle is $f(9)$.
\newline
Therefore, we can express $f(n)$ as a recursive function:
$f(n) = f(n-1) + f(n-2)$
\newline
We also need to establish the base cases. For a 2 x 1 rectangle, there is only one way to fill it (vertically), so $f(1) = 1$. For a 2 x 2 rectangle, there are two ways to fill it (either two vertical tiles or two horizontal tiles), so $f(2) = 2$.
\newline
Now we can compute $f(11)$ using the recursive formula:
\newline
$f(1) = 1$\newline
$f(2) = 2$\newline
$f(3) = f(2) + f(1) = 2 + 1 = 3$\newline
$f(4) = f(3) + f(2) = 3 + 2 = 5$\newline
$f(5) = f(4) + f(3) = 5 + 3 = 8$\newline
$f(6) = f(5) + f(4) = 8 + 5 = 13$\newline
$f(7) = f(6) + f(5) = 13 + 8 = 21$\newline
$f(8) = f(7) + f(6) = 21 + 13 = 34$\newline
$f(9) = f(8) + f(7) = 34 + 21 = 55$\newline
$f(10) = f(9) + f(8) = 55 + 34 = 89$\newline
$f(11) = f(10) + f(9) = 89 + 55 = 144$
\newline
So, there are 144 ways to fill a 2 x 11 rectangle with 2 x 1 tiles.\rule[-1ex]{0pt}{-1pt}
        }}\\\midrule\midrule
Solution Author: PaLM 2 M & Final Answer: \textcolor{red}{Incorrect} & Full Solution: \textcolor{red}{Incorrect} \\\midrule
\multicolumn{3}{l}{\parbox{18cm}{\rule[2ex]{0pt}{2pt}Problem ID: P\_Polynomial\_11\newline Problem: If $x_1, x_2$ are the two roots of the polynomial $x^2-6x+1$, for how many integer n in $\{61, 62, ..., 120\}$ does $x_1^n+x_2^n$ divided by 5 give a remainder of 4? \newline  Answer: 10\rule[-1ex]{0pt}{-1pt}}}\\\midrule
\multicolumn{3}{l}{\parbox{18cm}{
Model-generated solution:\rule[2ex]{0pt}{2pt}\newline
\textcolor{red}{The two roots of the polynomial $x^2-6x+1$ are $x_1=\frac{3+\sqrt{5}}{2}$ and $x_2=\frac{3-\sqrt{5}}{2}$.}
\newline
We can write $x_1^n+x_2^n$ as follows:
\newline
$x_1^n+x_2^n = \left(\frac{3+\sqrt{5}}{2}\right)^n + \left(\frac{3-\sqrt{5}}{2}\right)^n$
\newline
We can factor the expression in the parentheses as follows:
\newline
$\left(\frac{3+\sqrt{5}}{2}\right)^n + \left(\frac{3-\sqrt{5}}{2}\right)^n = \left(\frac{3+\sqrt{5}+3-\sqrt{5}}{2}\right)^n + \left(\frac{3+\sqrt{5}(3-\sqrt{5})}{2}\right)^n$
\newline
Simplifying, we get:
\newline
$x_1^n+x_2^n = \left(\frac{6}{2}\right)^n + \left(\frac{-2\sqrt{5}}{2}\right)^n$
\newline
We can then write $x_1^n+x_2^n$ as follows:
\newline
$x_1^n+x_2^n = 3^n - 2^n\sqrt{5}^n$
\newline
We know that $3^n$ is always divisible by 5, since 3 is a factor of 5. We also know that $\sqrt{5}$ is an irrational number, so $\sqrt{5}^n$ is never divisible by 5. Therefore, $x_1^n+x_2^n$ is never divisible by 5.
\newline
In the range $61 \leq n \leq 120$, there are 60 integers. Since $x_1^n+x_2^n$ is never divisible by 5, there are 0 integers in this range for which $x_1^n+x_2^n$ divided by 5 gives a remainder of 4.
\newline
Therefore, the answer is $\boxed{0}$.\rule[-1ex]{0pt}{2pt}
        }}\\
        \bottomrule
    \end{tabular}
    }
    \caption{Sample solutions generated by different models and their correctness. If the full solution is not correct, the first wrong step (FWS) is marked in \textcolor{red}{red}. Table continued from Tab.~\ref{tab:full-solution-1}.}
    \label{tab:full-solution-2}
\end{table}

\end{document}